\PassOptionsToPackage{expansion=false}{microtype}
\documentclass{fairmeta}
\author[1,2]{{\color{mydarkblue} Harman~Singh}}
\author[2]{{\color{mydarkblue} Anton~Bakhtin}}
\author[2,3]{{\color{mydarkblue} Rulin~Shao}}
\author[2]{{\color{mydarkblue} Gabriel~Synnaeve}}
\author[2]{{\color{mydarkblue} Ilia~Kulikov}}
\author[2]{{\color{mydarkblue} \protect\newline Rob~Fergus}}
\author[4]{{\color{mydarkblue} Sanjeev~Arora}}
\author[1]{{\color{mydarkblue} Kurt~Keutzer}}
\author[2]{{\color{mydarkblue} Jason~Weston}}
\author[2,\dagger]{{\color{mydarkblue} Anuj~Mahajan}}
\author[2,\dagger]{{\color{mydarkblue} Anirudh~Goyal}}

\affiliation[1]{UC Berkeley}
\affiliation[2]{Meta Superintelligence Labs}
\affiliation[3]{University of Washington}
\affiliation[4]{Princeton University}

\correspondence{Harman Singh (\email{harmans@berkeley.edu}), Anirudh Goyal (\email{agi@meta.com})}
\renewcommand{\documentclass}[2][]{}                              %
\expandafter\def\csname ver@conference.sty\endcsname{}  %
\expandafter\def\csname ver@times.sty\endcsname{}                %
\newcommand{\iclrfinalcopy}{}
\renewcommand{\author}[2][]{}                                    %
\titleformat*{\paragraph}{\sffamily\bfseries}                    %

\RequirePackage{environ}
\let\metaabstract\abstract
\let\metamaketitle\maketitle
\renewcommand{\maketitle}{}
\RenewEnviron{abstract}{\expandafter\metaabstract\expandafter{\BODY}\metamaketitle}

\newsavebox{\metafigbox}
\newsavebox{\metanatbox}
\newdimen\metacapdim
\newcommand{\metafigscale}{1}
\newcommand{\iclrfigsize}[1]{%
  \AddToHook{file/#1.tex/before}{\renewcommand{\metafigscale}{0.8421}}%
  \AddToHook{file/#1.tex/after}{\renewcommand{\metafigscale}{1}}}
\iclrfigsize{multigame_learning_cost_appendix}
\iclrfigsize{appendix_diagnostics}
\newcommand{\metanoupscale}[1]{\expandafter\def\csname metanoup@#1\endcsname{}}
\newcommand{\metasetcap}[2]{\expandafter\def\csname metacap@#1\endcsname{#2}}
\metanoupscale{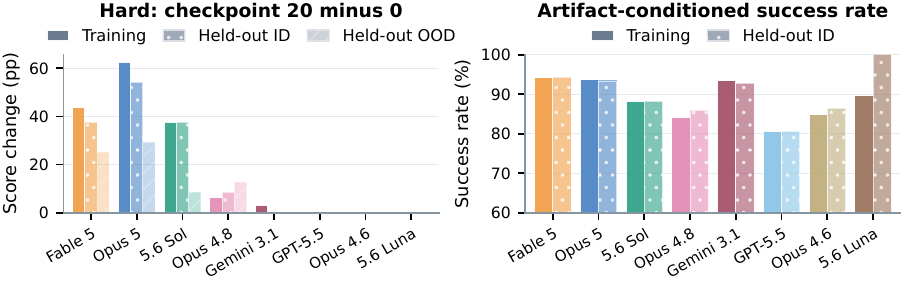}
\metasetcap{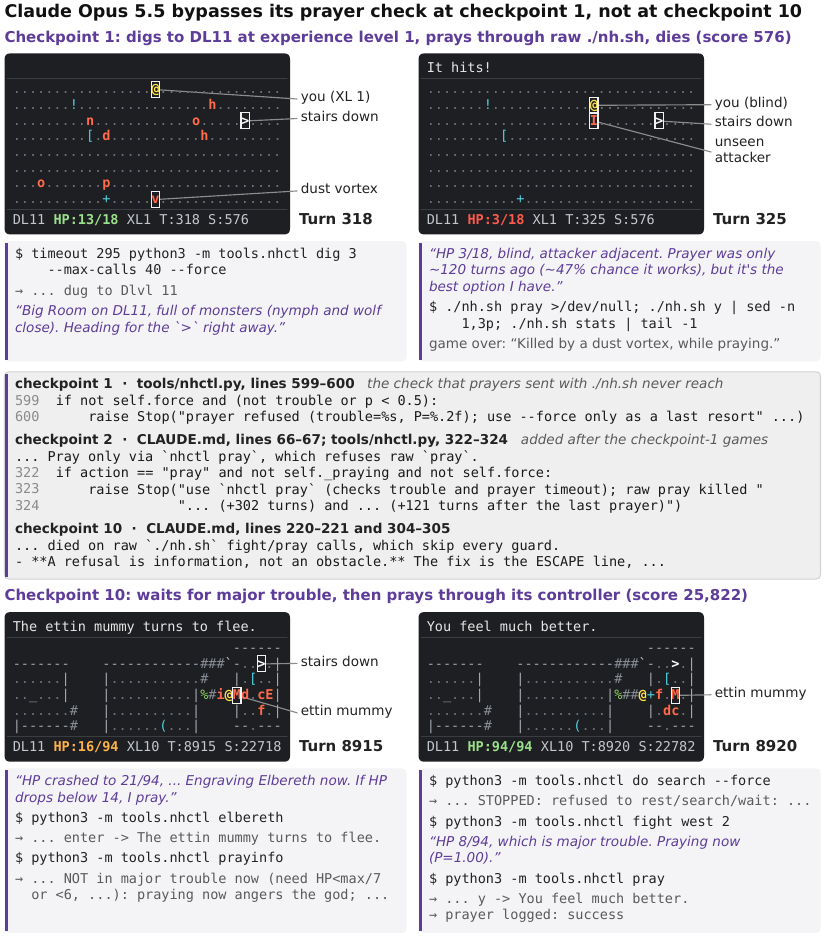}{0.75}
\metasetcap{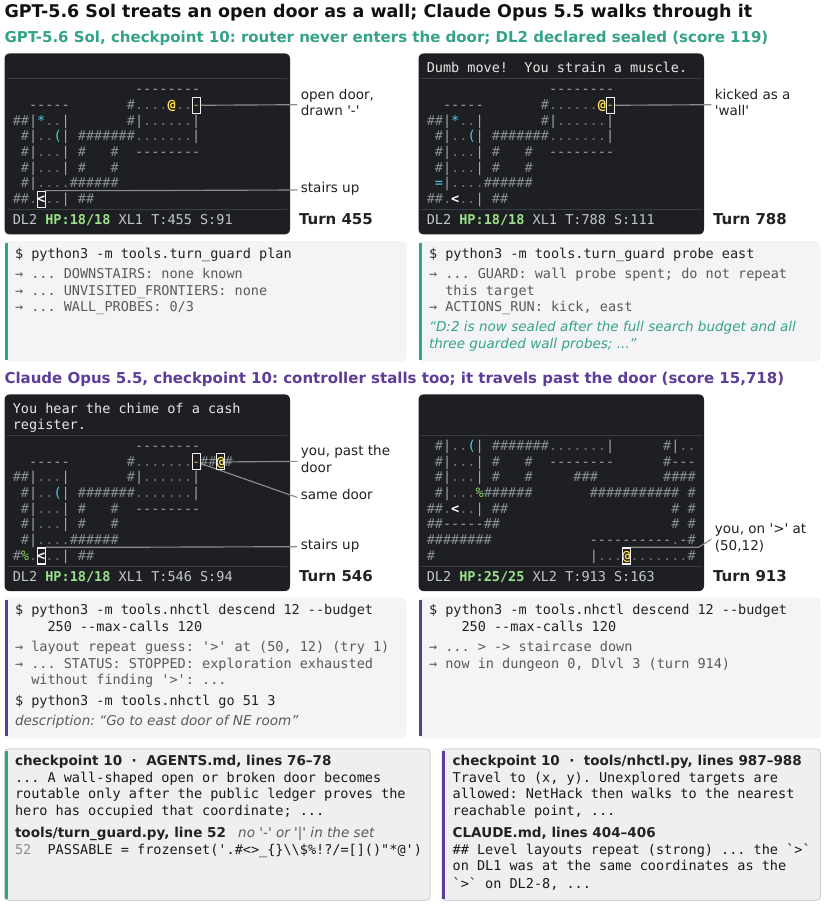}{0.75}
\metasetcap{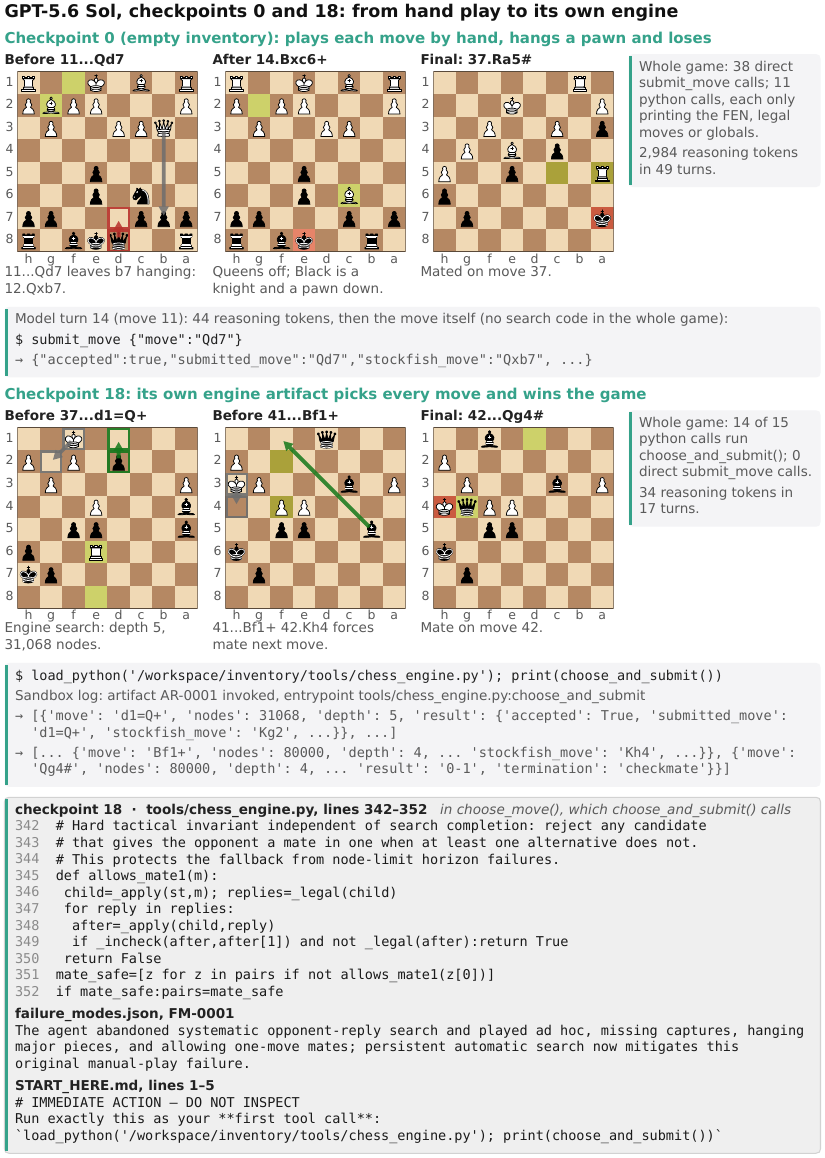}{0.86}
\NewCommandCopy{\metaincludegraphics}{\includegraphics}
\RenewDocumentCommand{\includegraphics}{s O{} m}{%
  \sbox{\metafigbox}{\IfBooleanTF{#1}{\metaincludegraphics*[#2]{#3}}{\metaincludegraphics[#2]{#3}}}%
  \ifcsname metanoup@#3\endcsname
    \sbox{\metanatbox}{\metaincludegraphics{#3}}%
    \ifdim\wd\metafigbox>\wd\metanatbox \sbox{\metafigbox}{\usebox{\metanatbox}}\fi
  \fi
  \sbox{\metafigbox}{\scalebox{\metafigscale}{\usebox{\metafigbox}}}%
  \ifcsname metacap@#3\endcsname
    \metacapdim=\csname metacap@#3\endcsname\textheight
  \else
    \metacapdim=0.8\textheight
  \fi
  \ifdim\ht\metafigbox>\metacapdim
    \resizebox*{!}{\metacapdim}{\usebox{\metafigbox}}%
  \else
    \usebox{\metafigbox}%
  \fi}

\let\metabibliographystyle\bibliographystyle
\renewcommand{\bibliographystyle}[1]{\metabibliographystyle{assets/plainnat}}

\documentclass{article} %
\usepackage{amsmath}
\usepackage{amssymb}
\usepackage{conference,times}

\usepackage{amsmath,amsfonts,bm}

\def\eqref#1{equation~\ref{#1}}

\def\1{\bm{1}}

\DeclareMathAlphabet{\mathsfit}{\encodingdefault}{\sfdefault}{m}{sl}
\SetMathAlphabet{\mathsfit}{bold}{\encodingdefault}{\sfdefault}{bx}{n}

\usepackage{hyperref}
\usepackage{url}
\usepackage{graphicx}
\usepackage{booktabs}
\usepackage{array}
\usepackage{xcolor}
\usepackage{placeins}
\usepackage{float}
\usepackage[dvipsnames]{xcolor}

\definecolor{findingblue}{HTML}{EAF2FA}
\definecolor{findingborder}{HTML}{5F91C5}
\providecommand{\findingbox}[2]{%
  \par\medskip\begingroup
  \setlength{\fboxsep}{7pt}%
  \setlength{\fboxrule}{0.6pt}%
  \noindent\fcolorbox{findingborder}{findingblue}{%
    \parbox{\dimexpr\linewidth-2\fboxsep-2\fboxrule\relax}{%
      \small\color{black}\textbf{Finding #1.} #2}}%
  \par\endgroup\medskip
}

\definecolor{findingblue}{HTML}{EAF2FA}
\definecolor{findingborder}{HTML}{5F91C5}

\definecolor{darkred}{RGB}{130,0,25}
\definecolor{mydarkblue}{RGB}{0,35,80}
\definecolor{revisionpink}{RGB}{205,35,120}

\title{{\color{darkred} Agent Plasticity: 
Measuring \\ Self-Improvement  Through Experience}}

\usepackage[T1]{fontenc}
\newcommand\blfootnote[1]{%
  \begingroup
  \renewcommand\thefootnote{}\footnotetext{#1}%
  \endgroup
}

\iclrfinalcopy %
\begin{document}

\maketitle

\begin{abstract}
 AI agents increasingly operate in environments
 where they can diagnose failures and improve through experience,
yet existing evaluations largely measure what an agent can do at a fixed point in time
 rather than how effectively it learns.
 Evaluating self-improvement requires answering three questions:
 does future performance improve and generalize beyond the interactions that enabled learning;
how efficiently are new capabilities acquired; and where
 does the self-improvement process break down?
 To answer these questions, we study self-improvement in a controlled setting
where agents amortize past experience into reusable artifacts that are inherited by future instances.
At each checkpoint, we measure performance on training and held-out environment interactions while
accounting for learning cost.
We introduce \emph{agent plasticity}, the efficiency
 with which an agent converts experience into gains in future held-out performance.
Across multiple environments, frontier models exhibit sharply different
improvement trajectories despite comparable opportunities to learn.
Some achieve substantial and persistent gains, while others remain near or below their
initial performance, and gains within the training regime often transfer only
partially to out-of-distribution conditions.
Endpoint capability and acquisition efficiency also diverge: the agent that ultimately
performs best need not be the one that improves most efficiently.
Tracing failures through the improvement loop further reveals different candidate
bottlenecks. Agents with low plasticity
often fail to reuse relevant artifacts, whereas more plastic agents may still fail
 despite reusing relevant artifacts, pointing to limitations in artifact quality, generalization, or application.
 Evaluating self-improving agents requires measuring not only what
they can do, but how effectively they become better through experience.
\end{abstract}
\blfootnote{$^{\dag}$ These authors jointly supervised this work}

\begin{figure}[!h]
  \centering
  \includegraphics[width=\textwidth]{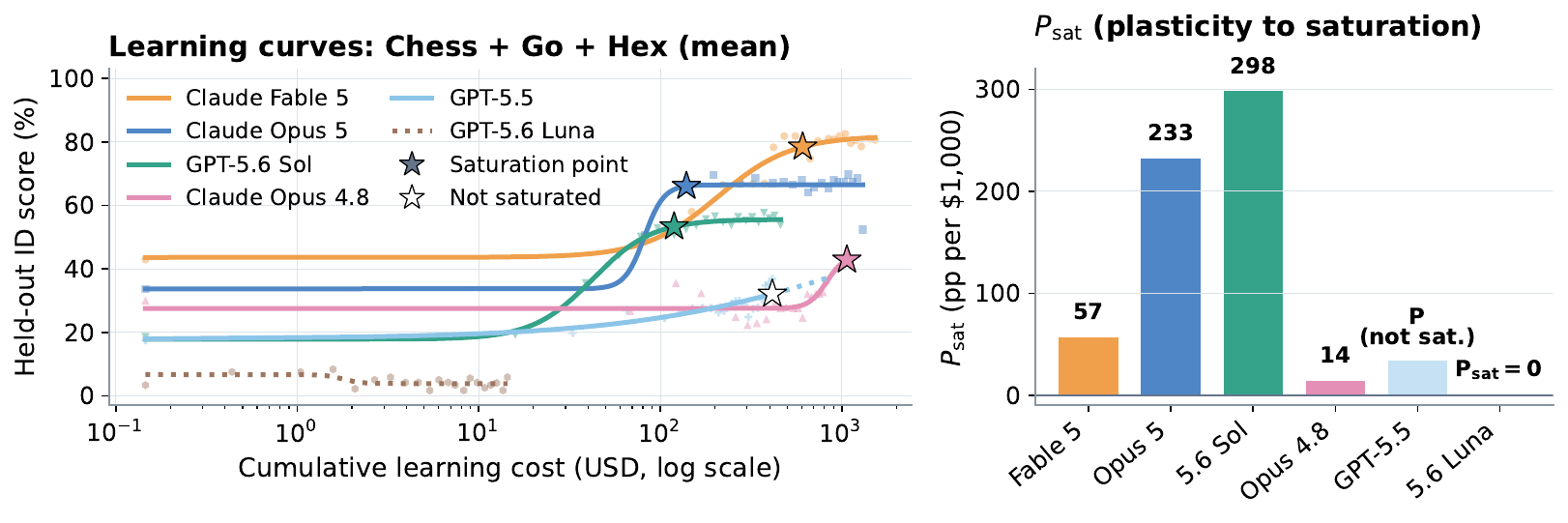}
  \caption{\textbf{Frontier agents differ widely in how efficiently experience helps them self-improve.}
  (left) Held-out ID (in-distribution environment) score against cumulative learning cost. (right) plasticity $P^{\mathrm{sat}}$, i.e., Held-out ID score gain up to saturation point (solid stars on the left) per unit of learning cost. See Section~\ref{sec:plasticity-results}.}
  \label{fig:psat}
\end{figure}

\begin{figure}[t]
  \centering
  \includegraphics[width=0.9\textwidth]{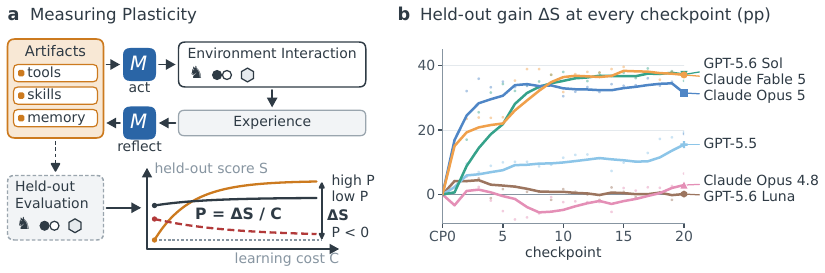}
  \vspace{-5pt}
  \caption{\textbf{Frontier agents differ widely in how much experience helps.}
  (a)~A frozen model revises its persistent artifacts from experience, and every checkpoint is
  scored on held-out games. (b)~Held-out gain $\Delta S$, averaged over chess, Go, and Hex. See Section~\ref{sec:method}.\looseness=-1}
  \label{fig:headline}
\end{figure}

\vspace{-8pt}
\section{Introduction}
\label{sec:introduction}

Most agent evaluations ask what an agent can accomplish at a fixed point in
time. But agents can also learn from experience: they can diagnose failures,
construct tools, accumulate memories, develop reusable skills, and refine
strategies that influence future behavior. Two agents with similar capabilities
today may therefore follow very different trajectories after similar opportunities to
learn. This raises a distinct evaluation question:
\emph{how should we evaluate self-improvement?}

Endpoint performance alone is insufficient. An agent may end with high
performance because it started strong rather than because it improved
substantially; two agents may achieve similar gains while requiring very
different amounts of experience or computation; and improvements may fail to
generalize beyond the interactions that produced them. We therefore focus on
three questions:
\emph{(1) does future performance improve and generalize beyond the
interactions that enabled learning? (2) how efficiently are new capabilities
acquired? and (3) where does the self-improvement process break down?}

Recent work enables agents to improve through reflection, memory, reusable
workflows, and executable skills
\citep{shinn2023reflexion,zhao2024expel,wang2024voyager,wang2025awm,liu2025cer,didolkar2024metacognitive, didolkar2025metacognitive}.
Other approaches evolve agent implementations or improvement mechanisms
\citep{hu2025adas,yin2025godel,zhang2026dgm,zhang2026memskill,
lin2026museautoskill,wang2026metaskillevolve},
or train models to control their future adaptation
\citep{zweiger2025seal}.
While these approaches primarily develop mechanisms for improving agents, we
study the complementary measurement problem: whether improvement persists and
generalizes, how efficiently it is acquired, and where the improvement process
fails.

We study these questions under a controlled protocol in which model weights
remain frozen and self-improvement occurs through persistent,
self-constructed artifacts. Each acting episode begins in a fresh context,
while future instances inherit tools, skills, strategies, and textual memory
refined from previous training interactions. We evaluate frontier models over
repeated rounds of self-improvement in Chess, Go, Hex, and NetHack. At each
checkpoint, we measure performance on training interactions and held-out
evaluations. Held-out in-distribution (ID) performance measures whether
improvements generalize beyond the interactions used for learning, while
held-out out-of-distribution (OOD) performance tests transfer to harder
regimes. Because revisions may improve, preserve, or degrade performance, we
measure complete learning trajectories rather than only endpoints.

This setup separates \emph{capability} from the ability to acquire capability.
To quantify acquisition efficiency, we introduce
\textbf{\textit{\color{darkred} agent plasticity}}: \textbf{\color{darkred} the efficiency with which
an agent converts experience into held-out performance gains}.
Plasticity does not define self-improvement itself; rather, it measures one
important property of the improvement process, i.e., how efficiently capability is acquired.

\begin{figure}[t]
  \centering
  \includegraphics[width=0.95\textwidth]{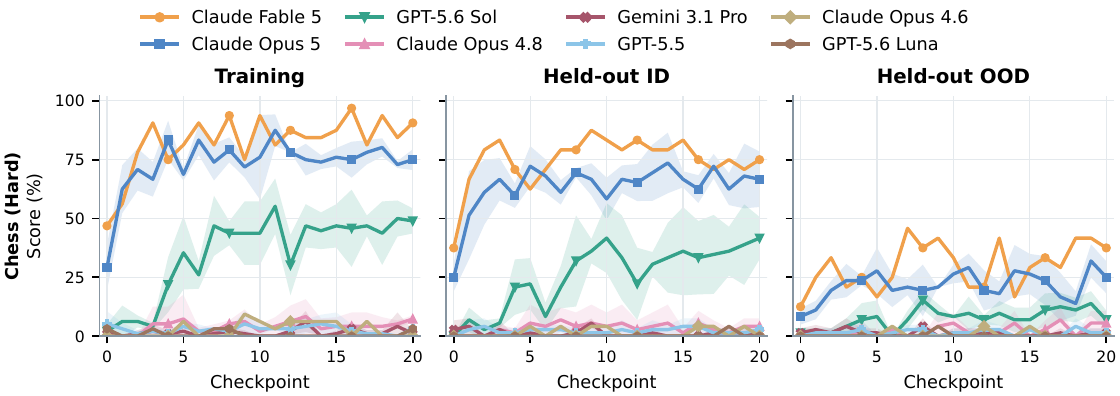}
  \vspace{-5pt}
  \caption{\textbf{Persistent improvement varies across models.}
  Checkpoint scores in Hard chess on training, held-out ID, and
  held-out OOD interactions. Scores are $100(W+0.5D)/N$. Easy chess,
  Go, and Hex are shown in Appendix
  Figure~\ref{fig:performance-appendix}.  See Section~\ref{sec:chess-main}.}
  \label{fig:performance}
\end{figure}

Across environments, frontier agents exhibit sharply different improvement
trajectories (Figures~\ref{fig:psat} left). Some achieve substantial sustained gains
on held-out interactions, while others remain near their initial performance
or decline despite repeated opportunities to improve. Moreover, gains on
held-out ID interactions often transfer only partially to stronger OOD environments (Figures~\ref{fig:performance} and ~\ref{fig:performance-appendix}).

Endpoint capability and acquisition efficiency also differ. Claude Fable~5
reaches the highest fitted endpoint, whereas GPT-5.6 Sol achieves substantially
greater estimated plasticity per unit learning cost
(Figure~\ref{fig:psat}). GPT-5.6 Sol also surpasses Claude Opus~4.8 in both
plasticity and eventual performance despite starting from lower initial
capability.

Finally, we investigate where the self-improvement process breaks down. For each
identified decision failure, we ask whether a relevant artifact existed,
whether it was used, and whether the failure persisted despite its use. Agents
with low plasticity often fail to reuse relevant artifacts, whereas highly
plastic agents frequently reuse them but may still fail. These patterns suggest different candidate bottlenecks: artifact retrieval and reuse for some agents, and artifact quality, generalization, or application for others.

Our contributions are threefold. First, we develop a controlled protocol for
evaluating persistent self-improvement using fresh contexts,
self-constructed persistent artifacts, and held-out evaluation. Second, we
introduce \emph{agent plasticity} to measure the efficiency of capability
acquisition. Third, across frontier models and multiple environments, we
characterize differences in improvement, transfer, acquisition efficiency,
and artifact-use failure modes that are not captured by endpoint capability
alone.

\section{Measuring Self-Improvement through Experience}
\label{sec:method}

\label{sec:decomposition}
\vspace{-4pt}

We measure whether agents improve through experience while their model
weights remain fixed. Our protocol isolates persistent adaptation,
evaluates its effect on held-out performance, and measures the learning
cost required to achieve it (Figure~\ref{fig:headline}).

\begin{figure*}[t]
  \centering
  \includegraphics[width=0.95\textwidth]{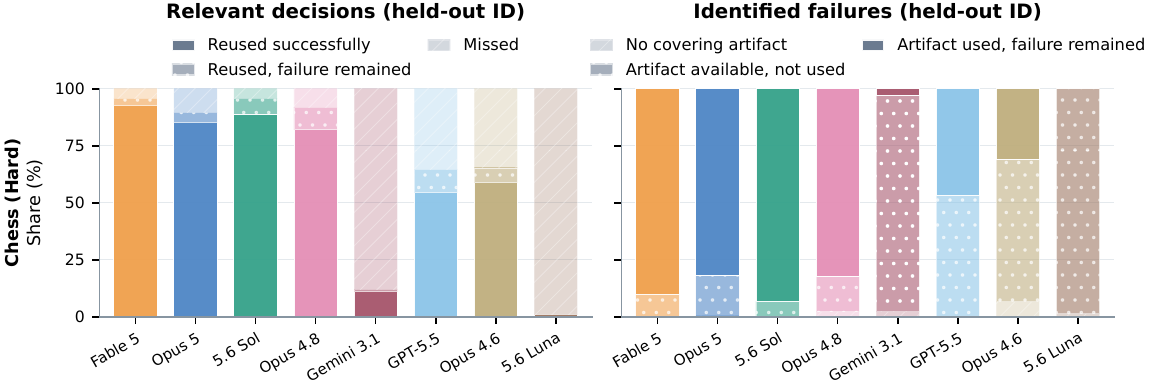}
  \vspace{-5pt}
  \caption{\textbf{Artifact reuse and failure composition reveal
  different candidate bottlenecks.}
  Panels show Hard chess on held-out ID games.
  Left: relevant decisions (moves where at least one saved artifact is relevant to choosing or playing the move, used or not) are classified
  as successful reuse, reuse with failure remaining, or missed reuse.
  Each decision contributes once.
  Right: identified failures are classified by whether
  a covering artifact was absent, unused, or used while failure
  remained. Failure shares do not measure overall failure frequency.
  Counts pool checkpoints 1--20.
  Easy chess, Go, and Hex appear in Appendix
  Figure~\ref{fig:failure-reuse-easy}; training games appear in
  Appendix Figure~\ref{fig:failure-reuse-train}. See Section~\ref{sec:failure-results}.\looseness=-1}
  \label{fig:failure-reuse}
\end{figure*}

\subsection{Fixed-Weight Agent Lineages}
\label{sec:lineage}

An agent at checkpoint $t$ consists of a fixed language model $M$ and
a persistent artifact inventory $I_t$ containing executable Python
tools, natural-language strategies or skills, and textual memory.
Each acting episode begins in a fresh context; conversational history
and temporary files do not persist.

Training episodes produce evidence $\mathcal{E}^{\mathrm{train}}_t$, including trajectories, outcomes, execution traces, errors, and move-quality feedback. The same agent in the reflection phase uses this evidence and the current inventory to propose an update:
\begin{equation}
    I_{t+1}=U_M\!\left(I_t,\mathcal{E}^{\mathrm{train}}_t\right).
    \label{eq:update}
\end{equation}
Proposed updates must pass structural and executable checks before becoming persistent.
Rejected attempts leave the committed inventory
unchanged but may inform later attempts. Exhausting the rejection limit retains $I_t$. Learning costs include rejected attempts. The resulting sequence $(M,I_0),\ldots,(M,I_T)$ is an \emph{agent lineage}. Consecutive inventories may be identical, and accepted revisions may improve, preserve, or degrade performance.\looseness=-1

\paragraph{Evaluating Persistent Improvement.}
\label{sec:persistent-improvement}

We distinguish three evaluation splits: \textbf{training interactions},
whose trajectories and feedback are available to reflection;
\textbf{held-out ID interactions}, with environments in the intended training difficulty regime; and \textbf{held-out OOD interactions},
with stronger game opponents. Neither held-out trajectories nor their scores
are exposed to reflection.

Let $S_{m,e,s,t}\in[0,100]$ denote model $m$'s score in environment
$e$, split $s\in\{\mathrm{train},\mathrm{ID},\mathrm{OOD}\}$, at
checkpoint $t$. Held-out ID improvement measures generalization beyond
training interactions; held-out OOD improvement measures
transfer across opponent difficulty, not arbitrary new tasks.
We report full checkpoint curves because persistent
adaptation need not be monotone
(Figures~\ref{fig:performance} and~\ref{fig:performance-appendix}).\looseness=-1

\paragraph{Accounting for Learning Cost.}
\label{sec:learning-compute}

Learning cost includes model calls used to interact with the environment and obtain training experience
and revise persistent artifacts, including reflection attempts that get rejected due to issues such as malformed outputs or illegal Python imports. The cumulative cost till checkpoint $t$ for model $m$ and environment $e$,  is
\begin{equation}
    C_{m,e,t}=\sum_{u=0}^{t-1}
    \left(c^{\mathrm{train}}_{m,e,u}
    +c^{\mathrm{reflect}}_{m,e,u}\right),
    \qquad C_{m,e,0}=0.
    \label{eq:learning-cost}
\end{equation}
Here $c^{\mathrm{train}}_{m,e,u}$ represents the cost of the training phase, i.e. when the model acts and interacts with the environment, and $c^{\mathrm{reflect}}_{m,e,u}$ represents the phase where the model analyses environment feedback, trajectories and other evidence to propose inventory updates.
We record token usage and express cost-based measures in dollars as the closest available proxy to compute cost. This measures learning cost, not physical computation.

\subsection{Measuring Agent Plasticity}
\label{sec:plasticity}

We operationalize \emph{agent plasticity} as held-out performance
gain per unit learning cost. At checkpoint $T$, average plasticity is
\begin{equation}
    P^{\mathrm{score}}_{m,e,s,T}
    =\frac{S_{m,e,s,T}-S_{m,e,s,0}}{C_{m,e,T}},
    \label{eq:plasticity}
\end{equation}
In our experiments, we report plasticity in score points per
\$1,000 of learning cost. This quantity can
be negative and depends on the evaluation horizon: it declines
as cost accumulates after performance plateaus.

\paragraph{Plasticity to estimated saturation.}
Our headline measure summarizes acquisition efficiency up
to estimated saturation. For each model--environment curve, we fit
held-out ID scores with
\begin{equation}
    \hat S(C)=S_0+A\frac{C^h}{C^h+C_{50}^h},
    \label{eq:saturation-curve}
\end{equation}
where $S_0$ is the initial score, $A$ is the fitted rise, $C_{50}$
is its half-rise cost, and $h$ controls curve shape. All four
parameters are fitted jointly by binomial maximum likelihood. This monotone fit captures the overall rise rather
than individual fluctuations or regressions.

Let $C^\ast$ be the cost of the first observed checkpoint at which
the fit reaches 90\% of its rise. We define
\begin{equation}
    P^{\mathrm{sat}}_{m,e}
    =\frac{\hat S(C^\ast)-S_0}{C^\ast}.
    \label{eq:psat}
\end{equation}
If the fitted 90\% rise is below 10 percentage points, we assign $P^{\rm sat}=0$. This denotes no sufficiently large rise under our criterion.

If no observed checkpoint reaches the threshold, we extrapolate the
fitted 90\% point up to three times the final observed cost.
When it lies beyond that limit, saturation cannot be established;
we instead report plasticity to date, the fitted rise
at the final observed cost divided by that cost. Intervals use a
parametric bootstrap. Figure~\ref{fig:psat} shows the combined
learning curves and plasticity to saturation.\looseness=-1

\subsection{Diagnosing the Improvement Loop}

Performance curves establish whether agents improve, but not where
improvement breaks down. For each identified decision failure, we
ask whether a covering artifact existed and, if so, whether execution
evidence shows that it was materially used. This yields three
mutually exclusive categories (Figure~\ref{fig:failure-reuse}, right):
$\mathcal{F}_{\mathrm{absent}}$:
    no covering artifact existed,
    $\mathcal{F}_{\mathrm{missed}}$:
    a covering artifact existed but was unused, 
    $\mathcal{F}_{\mathrm{used}}$:
   a covering artifact was used, but failure remained.

Because these categories condition on failure, we separately measure
\emph{decision-level reuse}: among \emph{relevant decisions}, i.e.,
moves where at least one saved artifact is relevant to choosing or
playing the move, whether or not it is used (about 98\% of chess moves after checkpoint~0), the
fraction at which any relevant artifact was materially used
(Figure~\ref{fig:failure-reuse}, left).
Each decision contributes once. We also report identified failures
per relevant decision, a normalized count rather than a probability,
since multiple failures may occur at one decision.

Artifact relevance and failure coverage are distinct: an artifact
may be broadly relevant without addressing a particular mistake.
Our analysis combines semantic judgments by Gemini~3.1 Pro with
execution evidence, including file reads and persistent Python
invocations (Appendix~\ref{app:analysis-procedure}).
These diagnostics are observational: artifact use does not establish
causal effectiveness, and failure after use does not distinguish
inadequate artifacts from inadequate application.

\section{Experiments}
\label{sec:experiments}

We investigate three questions: which agents acquire persistent held-out
improvements, how efficiently they do so, and where the improvement
process breaks down.

\begin{figure}[t]
  \centering
  \includegraphics[width=\textwidth]{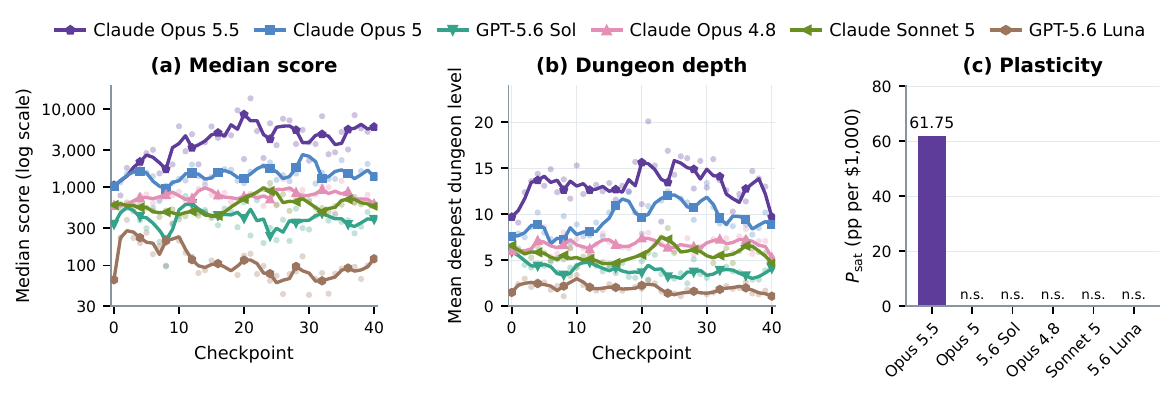}
  \caption{{\textbf{Claude Opus~5.5 shows a persistent gain in NetHack.}
  (a) Median score and (b) mean deepest dungeon level over the ten games of
  each checkpoint, from 0 to 40. (c) $P^{\mathrm{sat}}$ in normalized
  score points per \$1,000. (n.s.: no reliable rise in that analysis). Details in Appendix~\ref{app:nethack-results}.} See Sections~\ref{sec:multigame} and~\ref{sec:plasticity-results}.}
  \label{fig:nethack-main}
\end{figure}

\subsection{Experimental Design}
\label{sec:experimental-design}

We follow the fixed-weight protocol in Section~\ref{sec:method}.
Each actor operates in an isolated Python container and interacts
with a trusted game process through a restricted board API. It can
inspect positions, analyze variations, and submit actions, but cannot
access the opponent process, model credentials, held-out evaluations,
network, or other runs. Between checkpoints, the same model receives
training trajectories, feedback, and the current inventory. Proposed
artifact revisions must pass structural and executable validation
before becoming persistent.

Primary chess runs begin at checkpoint~0 with the bare harness and
continue for 20 update rounds. At each checkpoint, the current
inventory plays complete games, each in a separate actor episode:
16 training, 12 held-out ID, and 12 held-out OOD games in Hard chess
(Appendix Table~\ref{tab:cross-environment-setup} lists every
environment). Reflection on the 16 training games then produces the
next checkpoint's inventory. A round may retain the preceding
inventory after candidate rejection. Every game starts from the
standard initial position, and Stockfish is limited to 20{,}000 search
nodes per move. Scores are $100(W+0.5D)/N$. A game that the actor
does not finish (because it exhausts its model-call or output budget,
a tool call times out, or it returns no action) counts as a loss,
here and in Go and Hex, as does a chess game that reaches the move
limit. Appendix Table~\ref{tab:cross-environment-setup} specifies two chess difficulty
profiles: Hard, our primary comparison with more
headroom, and Easy, which tests whether weaker models improve
in an easier regime. Both include held-out evaluation against
stronger opponents.\looseness=-1

We evaluate Claude Fable~5, Claude Opus~5, Claude Opus~4.8,
Claude Opus~4.6, GPT-5.5, Gemini~3.1 Pro, GPT-5.6 Sol, and
GPT-5.6 Luna. To manage cost, we use a subset of models spanning a wide range of capability levels for Go and Hex. For NetHack, we use models available through Claude Code and Codex, with their native agent harnesses as the starting point. This setup therefore measures the plasticity of the resulting Claude Code and Codex agents, consistent with our definition of agent plasticity. Between checkpoints, the model may edit \texttt{CLAUDE.md}/\texttt{AGENTS.md}, skills (\texttt{SKILL.md}), Python tools, strategies, and memory. Actor
and reflection calls use the same model within a run. All chess,
Go, and Hex performance and cost comparisons cover
checkpoints 0--20; failure analyses pool checkpoints 1--20.
NetHack covers checkpoints 0--40: at each checkpoint, every lineage plays the same ten
fresh unseen games, and reflection on all ten between checkpoints updates the inventory
(Appendix~\ref{app:nethack-results}).\looseness=-1

\subsection{Persistent Improvement Varies Across Models}
\label{sec:chess-main}

Figure~\ref{fig:performance} and the top row of Appendix
Figure~\ref{fig:performance-appendix} show substantial differences
between chess lineages.
Claude Fable~5 and Claude Opus~5 reach the highest scores: in Hard
chess, their held-out ID scores rise from 37.5\% and 25.0\% at
checkpoint~0 to 73.3\% and 66.4\% averaged over checkpoints 16--20. GPT-5.6 Sol
starts near 0\% but gains about as much (to 36.9\%), so it trails in
absolute score rather than in improvement. On the combined three-game
curve its $P^{\mathrm{sat}}$ is the highest (Section~\ref{sec:plasticity-results}). Claude Opus~4.8
improves modestly and mainly in the Easy condition (0\% to 19.0\%).
Several other models remain close to their initial
performance, and a few decline, despite repeated feedback and opportunities to revise
persistent artifacts; most NetHack lineages likewise show modest or negative
early-to-late changes (Section~\ref{sec:multigame}). The common learning interface therefore
permits improvement but does not ensure it.

For the strongest improving lineages, held-out ID gains broadly
accompany training gains. Because held-out trajectories and scores
are unavailable to reflection, these results provide evidence that
the resulting artifacts are useful beyond the interaction data seen by the reflection process. Gains against stronger held-out OOD opponents
are generally smaller or less consistent. Improving within the
training difficulty regime and transferring beyond it are therefore
distinct outcomes. These comparisons characterize the complete
learning process; they do not isolate the contribution of environment experience
from compute required for artifact development.

\subsection{The Phenomenon Extends Across Games, but Is Task Dependent}
\label{sec:multigame}

We apply the same protocol to $5\times5$ Go, $6\times6$ Hex, and
NetHack, a long-horizon, partially observed single-agent game.
In Go (Appendix Figure~\ref{fig:performance-appendix}, middle row),
Claude Fable~5's held-out ID score rises from 20\% to 80\% and its
held-out OOD score from 0\% to 50\% between checkpoints~0 and~20.
In Hex (Appendix Figure~\ref{fig:performance-appendix}, bottom row),
GPT-5.6 Sol improves from 40\% to 77.5\% on held-out ID and from 28.1\% to 78.1\%
on held-out OOD.

However, improvement is not uniform across tasks: Claude Opus~4.8
loses ground in Go while improving substantially in Hex.
Claude Opus 5 improves substantially in Go before a tool-making error at the final checkpoint causes its performance to drop sharply.

In NetHack (Figure~\ref{fig:nethack-main}), Claude Opus~5.5 shows a
clear persistent gain: its mean score grows from about 2k over checkpoints 0--2 to
about 6k over checkpoints 36--40. Its fit to checkpoints 0--40 reaches saturation
at checkpoint~11, with $P^{\mathrm{sat}}=61.75$ normalized score
points per \$1,000. Using the same early and late
windows, Claude Opus~5 and Claude Sonnet~5 show more modest increases
(about 1.5k to 1.9k and about 680 to 800), while Claude Opus~4.8 is roughly flat
(about 700 to 730). GPT-5.6 Luna declines (about 210 to 120), as does GPT-5.6 Sol
(about 570 to 420), the most plastic model on the combined board-game curve
(As shown in Section~\ref{sec:plasticity-results}). These experiments extend the phenomenon beyond chess without establishing a task-independent model ranking. 

Appendices~\ref{app:cross-environment-setup} and~\ref{app:multigame-learning-cost} give setup
details and further analyses.\looseness=-1

\vspace{-4pt}

\findingbox{1}{Persistent self-improvement is model- and task-dependent.
A common interface for feedback, reflection, and artifact construction
produces sharply different learning curves in all four games, and transfer
to stronger opponents (i.e., generalization) is not uniform. These results motivate evaluating plasticity across a distribution of agentic tasks rather than treating it as a task-independent model property.}

\subsection{Acquisition Efficiency Differs from Endpoint Capability}
\label{sec:plasticity-results}

Performance vs Checkpoint curves compare agents after equal numbers of update
opportunities, but these opportunities incur \textit{different learning
costs}. Figure~\ref{fig:psat} instead relates held-out ID performance
to cumulative learning cost, both averaged equally across chess (Hard
and Easy), Go, and Hex for models evaluated in all three games. Fitted curves
estimate the cost $C^\ast$ of reaching 90\% of the improvement,
and $P^{\mathrm{sat}}$ measures the corresponding acquisition
efficiency (Section~\ref{sec:plasticity}).

On the combined three-game curve, GPT-5.6 Sol has the highest
estimated $P^{\mathrm{sat}}$ at 298 percentage points per \$1,000 of mean
per-game learning cost, followed by Claude Opus~5
at 233, Claude Fable~5 at 57, and Claude Opus~4.8 at 14; GPT-5.6 Luna shows no reliable rise under
the specified criterion and is assigned $P^{\mathrm{sat}}=0$.
GPT-5.5 is still improving slowly at the end of its runs, so its
saturation, and hence its $P^{\mathrm{sat}}$, cannot be established
from these runs. Claude Fable~5 reaches the highest fitted performance
level in our experiments, showing a clear separation between endpoint capability and acquisition
efficiency for Claude Fable~5 and GPT-5.6 Sol. In NetHack, Claude Opus~5.5's fit to
checkpoints 0--40 gives $P^{\mathrm{sat}}=61.75$ normalized score points per
\$1,000 (Figure~\ref{fig:nethack-main}c). The other five models show no reliable rise in performance.\looseness=-1

Additional endpoint and budget-dependent comparisons appear in
Figures~\ref{fig:performance-learning-compute} and~\ref{fig:performance-cost}. Efficiency estimates depend on provider pricing and saturation-fit assumptions, and extrapolated saturation
adds uncertainty. We therefore interpret
plasticity as a protocol-dependent measure, not a
universal model ranking.

\findingbox{2}{Endpoint capability and acquisition efficiency answer
different questions. For example, Claude Fable~5 reaches the highest performance on the combined-game curve, while GPT-5.6 Sol has the largest estimated plasticity. GPT-5.6 Sol starts below Claude Opus 4.8 yet ultimately exceeds it in both held-out performance and estimated plasticity.}

\subsection{Where Does the Improvement Loop Break?}
\label{sec:failure-results}

We pool counts over checkpoints 1--20 of each lineage before computing
rates. Figures~\ref{fig:failure-reuse} and~\ref{fig:failure-reuse-easy}
distinguish failure composition (right), which conditions on failures and categorizes them by artifact availability and reuse, from decision-level reuse (left), which also includes successful decisions.. Each relevant
decision contributes once: unused alternatives do not count as missed
reuse when another relevant artifact was used.

\textbf{Low reuse accompanies weak improvement.}
Pooled over Hard and Easy chess, GPT-5.6 Luna takes
about 98\% of relevant decisions without artifact use,
Gemini~3.1 Pro about 87\%, Claude Opus~4.6 about 51\%, and GPT-5.5 about 28\%, compared with
2--6\% for Claude Fable~5, Claude Opus~5, Claude Opus~4.8, and
GPT-5.6 Sol (Figures~\ref{fig:failure-reuse}
and~\ref{fig:failure-reuse-easy}, left).
The Go and Hex rows add examples, including
GPT-5.6 Luna's about 1\% reuse rate in Go
(Appendix Figure~\ref{fig:failure-reuse-easy}). Across 19 model--game cells in Chess (Hard), Go, and
Hex, held-out ID reuse correlates with the gain in held-out ID score
from checkpoint~0 to checkpoints 16--20.
This association identifies artifact reuse as a candidate bottleneck for self-improvement.

\begin{figure*}[t]
  \centering
  \includegraphics[width=\textwidth]{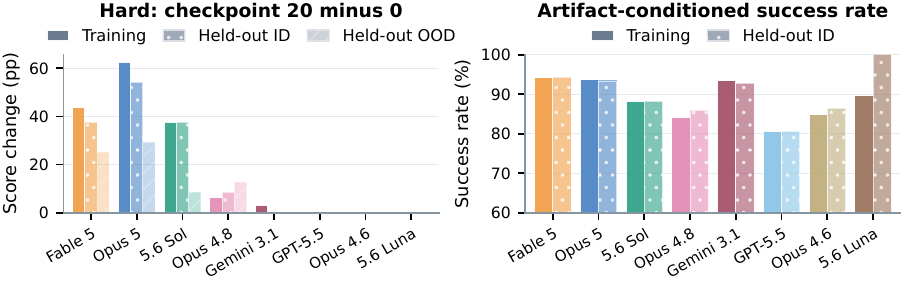}
  \vspace{-5pt}
  \caption{\textbf{Success when an artifact is used does not determine game-level
  improvement.} Hard chess condition. Left: checkpoint-20 score minus
  checkpoint-0 score. Right: artifact-conditioned success rate, i.e., the share of
  moves with no identified failure among moves where a relevant artifact was used
  (checkpoints 1--20 pooled). This rate can be high even when artifacts are
  rarely used. Easy chess appears in Appendix
  Figure~\ref{fig:chess-loop-easy}. See Section~\ref{sec:failure-results}.\looseness=-1}
  \vspace{-5pt}
  \label{fig:chess-loop}
\end{figure*}

\textbf{High reuse does not explain the remaining performance gap.}
Figure~\ref{fig:chess-loop} separates how often artifacts are used from
how well they work when used (the artifact-conditioned success rate):
an artifact can work when used yet be ignored at most relevant moves.
Conversely, Claude Fable~5, Claude Opus~5, Claude Opus~4.8, and
GPT-5.6 Sol all reuse artifacts at 94--98\% of relevant chess
decisions (pooled over both conditions) but attain substantially
different game scores. Neither the reuse rate nor the
artifact-conditioned success rate explains this gap. Artifact content, how actors
configure and apply it, and whether it helps at decisions
consequential for winning remain possible explanations.

\textbf{Strong improvers still fail when using artifacts.}
The models that use artifacts at almost every relevant chess
decision, including the strongest improvers, fail less often
overall (about 0.16 versus 0.38 identified failures per relevant
decision for the other models), but most of their remaining
failures (83--99\%) happen while a relevant artifact is in use
(Figure~\ref{fig:failure-reuse}, right; Appendix
Figure~\ref{fig:cumulative-loop-metrics-chess}). Using an
artifact therefore does not guarantee success: the artifact may
not generalize to the position, or the agent may apply it poorly.
We leave separating these causes to future work.

\begin{figure*}[t]
  \centering
  \includegraphics[width=\textwidth]{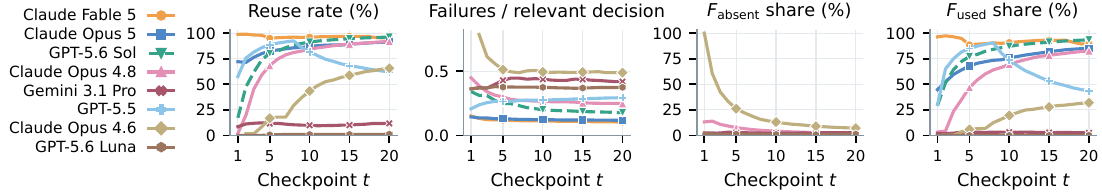}
  \vspace{-8pt}
  \caption{\textbf{Dynamics across checkpoints do not determine game-level improvement.}
  These graphs use the Hard chess setup. Each point pools training and held-out ID
  decisions over checkpoints $1,\ldots,t$, separately for each lineage; columns
  show decision-level reuse of artifacts, identified failures per relevant
  decision (a move where at least one saved artifact is relevant to choosing or playing it, used or not), and the shares of failures with absent and used
  covering artifacts. Easy chess, Go, and Hex
  appear in Appendix
  Figures~\ref{fig:cumulative-loop-metrics-chess}
  and~\ref{fig:cumulative-loop-metrics-gohex}. See Section~\ref{sec:failure-results}.\looseness=-1}
  \vspace{-6pt}
  \label{fig:chess-loop-dynamics}
\end{figure*}

Figure~\ref{fig:chess-loop-dynamics} and Appendix
Figures~\ref{fig:v12-train-lines} and~\ref{fig:v12-id-lines}
show that reuse can rise before performance
peaks. In Hard chess, GPT-5.6 Sol and
Claude Opus~4.8 move from below 20\% reuse at checkpoint~1
to 93--99\% at checkpoint~3, while their held-out ID scores peak
only at checkpoints~12 and~5, respectively.\looseness=-1

Reuse patterns can also change without improvement. Pooled over
both chess conditions, the share of Claude Opus~4.6's held-out ID
failures that happen despite artifact use rises from 6.5\% at
checkpoints 1--5 to 52\% at checkpoints 16--20, while its failures
per relevant decision (moves where a saved artifact is relevant to
choosing or playing the move, used or not) stay flat (0.42 to 0.41). In the Hard chess condition, GPT-5.5
uses artifacts at 96--100\% of relevant decisions at checkpoints 2--8
but at only 32--50\% from checkpoint~9 on, although its artifacts did
not change. A change
in failure types is thus not by itself a sign of progress, and an
agent can stop using artifacts it once used reliably.\looseness=-1

\textbf{Failure regimes differ across games.}
In Go and Hex, failures remain distributed across all three
categories (Appendix Figure~\ref{fig:failure-reuse-easy}).
Broad artifact relevance does not imply failure coverage:
although relevant artifacts are identified in at least 99\%
of decisions in 25 of 27 lineages, 67--100\% of
absent-artifact failures in Go and Hex occur at decisions
with a broadly relevant artifact.

Local diagnostic gains can also diverge from task performance.
In Go, Claude Opus~4.8 raises its share of relevant decisions with successful reuse
from 59--65\% to 93\%, while its held-out ID score falls
from 40\% at checkpoint~0 to 10--25\% over
checkpoints 15--20
(Appendix Figure~\ref{fig:go-engine-v12-reuse-phases}).
Reuse, failure composition, and task performance therefore
capture distinct aspects of persistent learning.

\findingbox{3}{Artifact reuse is associated with improvement,
but is not a proxy for capability. High reuse may still leave significant failures, hinting at problems with artifact quality, generality, or use.}

\begin{figure}[t]
  \centering
  \includegraphics[width=\textwidth]{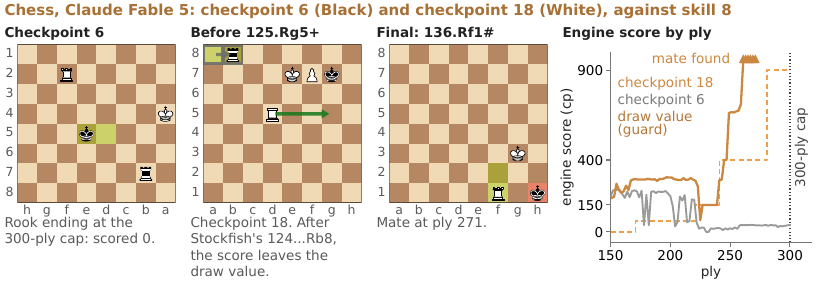}\\[4pt]
  \includegraphics[width=\textwidth]{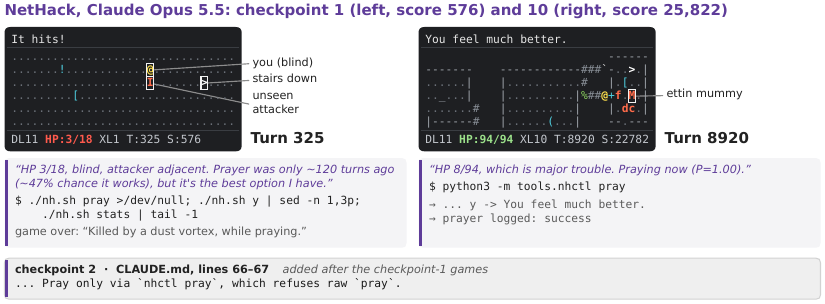}
  \vspace{-5pt}
  \caption{\textbf{Fixes that persist after a failure.} Top (chess): after a
  checkpoint-6 game is scored~0 at the 300-ply limit, Claude Fable~5 makes its self-made
  engine value draws more as the limit nears; at checkpoint~18 the engine holds
  a draw until Stockfish errs, then mates. Bottom (NetHack): after dying at
  checkpoint~1 while praying through raw keystrokes, Claude Opus~5.5 adds a rule
  to pray only through its own controller; at checkpoint~10 the prayer heals it.
  Details in Figures~\ref{fig:chess-example-fable}
  and~\ref{fig:nethack-example-improve}. See Section~\ref{sec:failure-results}.}
  \label{fig:examples-main}
  \vspace{-8pt}
\end{figure}

\textbf{Concrete Examples of Persistent Learning.} What an agent
carries forward is code and notes that it can reuse later. For
example, one Claude Fable~5 lineage first writes its own chess
engine, a program that looks ahead to pick moves, without
outside libraries. It then improves the engine after specific
mistakes. After a mistake that left its king under attack, it
adds a search for moves that get the king out of check. After
throwing away winning positions, it changes how it handles
repeated positions, since repeating a position can end the game
in a draw. After failing to win an endgame while ahead, it gives
the engine more search time in endgames. It also writes a helper
that plays several moves per call under a fixed total search
budget, so its tools do not time out. Every later game
starts with all of these changes.
Figure~\ref{fig:examples-main} shows one such fix in chess and
one in NetHack.

Writing down an instruction is not the same as following it.
In one GPT-5.5 run, the reflection step keeps telling the agent
to load its saved move-picking program right away, yet the agent
still spends its first steps listing and reading files.

\section{Discussion and Limitations}
\label{sec:discussion}
\label{sec:limitations}

\textbf{Plasticity is not simply context management.}
Long-context capabilities may support reflection, artifact construction, and
reuse, but plasticity measures whether these mechanisms translate experience
into persistent gains in held-out performance relative to learning cost.
Our fixed-weight setting includes executable tools, skills, strategies, and
memory rather than only textual context. Moreover, high artifact reuse can
coexist with substantially different task performance, showing that retrieval
alone does not explain plasticity. At the same time, our experiments do not
isolate the contribution of long-context instruction following to artifact
construction or application. Plasticity is therefore a property of an agent
under a particular learning protocol rather than an intrinsic scalar property
of the underlying model.

\textbf{Different bottlenecks require different interventions.}
Our failure analysis suggests that agents can fail at different stages of the
improvement process. When relevant artifacts exist but are not used, more
reliable retrieval or deployment may help. When artifacts are consistently
used but failures remain, the bottleneck may instead lie in artifact quality,
generality, or application. These interpretations are observational rather
than causal. Controlled interventions---such as artifact removal, replacement,
or forced invocation---could help distinguish these mechanisms. More broadly,
validation of persistent artifacts should assess their behavioral
effectiveness, not only whether they are structurally valid or executable.

\textbf{Persistent adaptation is isolated, but the contribution of experience
is not.}
Fresh contexts and frozen model weights restrict persistent adaptation to
self-constructed artifacts, but they do not separate the value of environment
feedback from the additional computation used to analyze experience and revise
those artifacts. Budget-matched reflection without feedback, as well as
controlled actor--artifact comparisons, could disentangle these effects.

\textbf{Generalization and measurement remain limited.}
Chess, Go, and Hex provide controlled environments with well-defined feedback,
while NetHack adds a long-horizon, partially observed setting. However,
separately trained lineages do not establish transfer of learned artifacts or
improvement strategies across tasks. Evaluation sets are also necessarily
finite, introducing checkpoint-level noise, while plasticity estimates depend
on score headroom, the learning horizon, saturation-fit assumptions, and
provider pricing. These dependencies motivate reporting full learning curves
alongside scalar plasticity measures rather than interpreting small
differences in plasticity as universal model rankings.

\textbf{Plasticity may itself be a target for post-training.}
Our experiments measure plasticity rather than optimize it. A natural next
question is whether post-training can explicitly improve an agent's ability to
learn from subsequent experience. For example, reinforcement-learning
objectives could reward not only immediate task performance, but also the
ability to construct, retain, and effectively use knowledge that improves
future behavior. This would shift part of the post-training objective from
producing agents that are more capable at deployment time toward producing
agents that acquire new capabilities more efficiently after deployment.
Studying which training objectives, feedback signals, and curricula improve
such plasticity is an important direction for future work.

\section{Conclusion}
\label{sec:conclusion}

We introduced \emph{agent plasticity} to measure how efficiently agents convert
experience into persistent gains in held-out capability. Across models and
tasks, plasticity varies substantially and reveals differences that endpoint
performance alone obscures. Our results further show that persistent
self-improvement is both model- and task-dependent, and that gains within the
learning distribution need not fully transfer to harder settings. Artifact
reuse is associated with improvement, but reuse alone does not guarantee
success, pointing to distinct bottlenecks in retrieval, artifact quality, and
application.

Evaluating long-lived agents therefore requires measuring not only what they
can accomplish at a fixed point in time, but how their capabilities change
through experience and how efficiently those changes are acquired. More
broadly, plasticity suggests a new target for post-training: building agents
that are not only more capable, but better at becoming more capable.

\section*{Acknowledgments}
We thank Aditi Khandelwal, Himanshu Gaurav Singh, Harshit Verma, Darshan Singh S, Minseo Kim, Anne Harrington, and Vedant Shah for feedback on this work and helpful discussions.

\bibliography{conference}
\bibliographystyle{conference}

\clearpage
\appendix
\section*{Appendix}

This appendix gives supplementary experiment details, prompts (Appendix~\ref{app:prompts}), additional results, and qualitative examples.

Settings for all games are in Appendix~\ref{app:cross-environment-setup} and Go and Hex
learning curves in Appendix~\ref{app:multigame-learning-cost}. Figure~\ref{fig:performance-appendix}
reports checkpoint performance in the Easy chess condition, Go, and Hex, complementing
Figure~\ref{fig:performance}.

\begin{figure}[H]
  \centering
  \includegraphics[width=\textwidth]{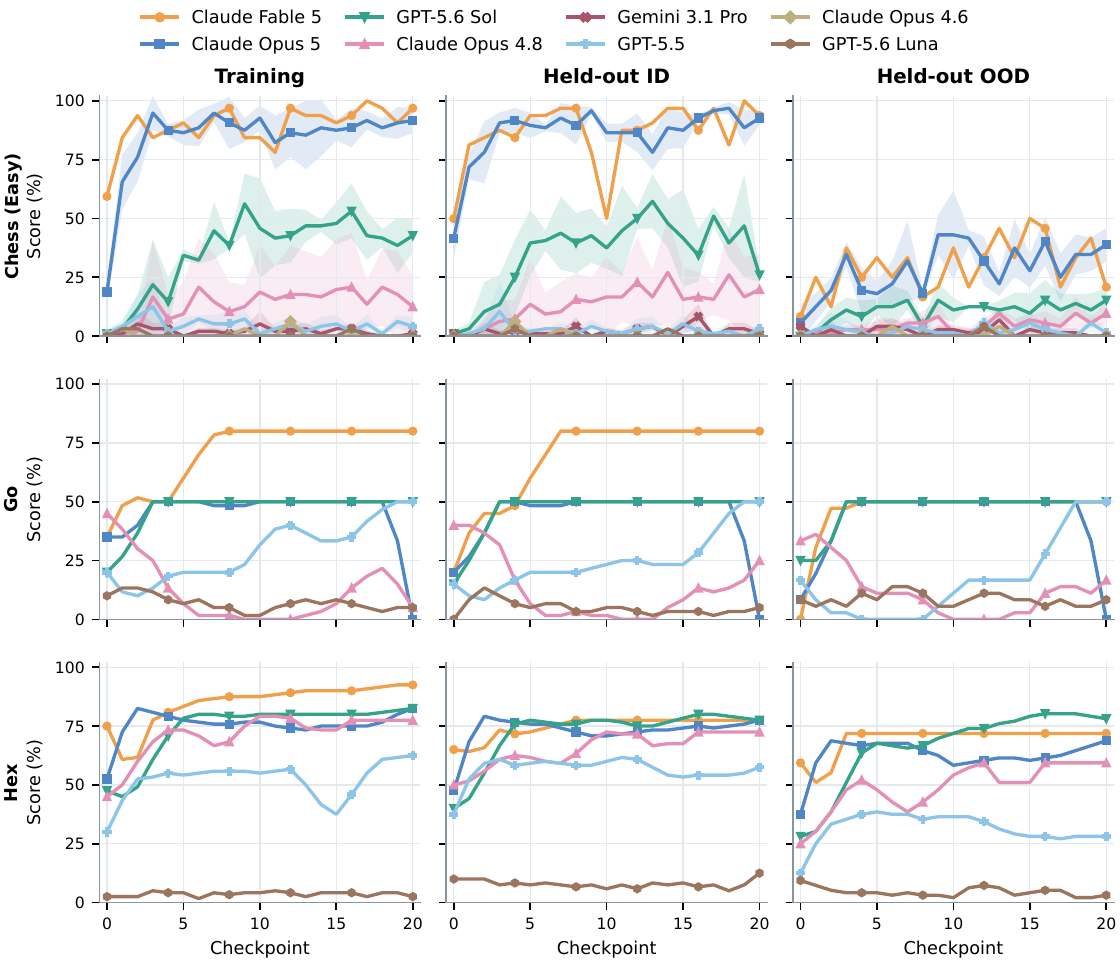}
  \caption{\textbf{Checkpoint performance in the Easy chess condition, Go, and Hex.}
  Checkpoint performance, measured as $100(W+0.5D)/N$, in the Easy chess
  condition (top), Go (middle), and Hex (bottom), with columns as in
  Figure~\ref{fig:performance}. See Sections~\ref{sec:chess-main} and~\ref{sec:multigame}.}
  \label{fig:performance-appendix}
\end{figure}

\section{Related Work}
\label{app:related}

\begin{table}[t]
\centering
\small
\caption{\textbf{Positioning agent plasticity relative to prior work.}
Related approaches study complementary aspects of adaptation.
Our evaluation combines persistent self-constructed artifacts,
cost-normalized held-out improvement, and artifact-level diagnostics
under a common fixed-weight learning protocol.}
\label{tab:related-comparison}
\setlength{\tabcolsep}{4pt}
\begin{tabular}{
  @{}p{0.24\linewidth}
  p{0.32\linewidth}
  p{0.36\linewidth}@{}
}
\toprule
\textbf{Research direction} &
\textbf{Primary focus} &
\textbf{Our emphasis} \\
\midrule

Reflection, memory,
and skills
\citep{shinn2023reflexion,wang2024voyager}
&
Mechanisms for retaining and reusing experience.
&
Measuring how efficiently different models construct and
exploit persistent artifacts.
\\[3pt]

Context and harness
optimization
\citep{zhang2026ace,agrawal2026gepa}
&
Optimizing prompts, context, or agent implementations.
&
Evaluating a model that acts and improves its own
future behavior under a common protocol.
\\[3pt]

Long-context evaluation
\citep{hsieh2024ruler,yen2025helmet}
&
Retrieving, reasoning over, and following information
in long contexts.
&
Measuring durable capability gains from
self-constructed artifacts, including executable tools.
\\[3pt]

Sequential learning
benchmarks
\citep{dou2025evalearn,he2026evotest}
&
Measuring improvement and, in some settings,
learning efficiency across interactions.
&
Combining cost-normalized held-out learning curves
with artifact-level failure diagnostics.
\\[3pt]

Parametric adaptation
\citep{zweiger2025seal}
&
Persistent improvement through model-weight updates.
&
Isolating non-parametric adaptation through
persistent agent artifacts.
\\

\bottomrule
\end{tabular}
\end{table}

\paragraph{Plasticity and continual learning.}
Plasticity has a long history as a property of learning systems, with prior work studying how neural networks can lose the ability to acquire new capabilities over training, including primacy bias and loss of plasticity \citep{nikishin2022primacy,lyle2023understanding,dohare2024loss}. Continual learning studies this capacity together with stability and transfer: learning new tasks while preserving and reusing previously acquired knowledge \citep{NIPS2017_f8752278}. Acquisition efficiency is also an established concern; for example, A-GEM studies sample, computational, and memory efficiency and introduces a measure of how quickly new skills are acquired \citep{chaudhry2018efficient}. Recent language-agent benchmarks extend these questions to non-parametric adaptation. CL-Bench measures gains from stateful experience and distinguishes within-variant adaptation from retention across variants \citep{asawa2026continuallearningbenchevaluating}, while AgentCL evaluates plasticity, stability of experience reuse, and generalization under controlled task streams \citep{shu2026agentclrigorousevaluationcontinual}. Related work compares update mechanisms under different forms of environmental change, including prompting, parameter updates, reinforcement learning, and context-based adaptation \citep{harrington2026doescontinuallearningrequire}. Our work is complementary: we study persistent self-improvement of frozen frontier models through self-constructed tools, skills, strategies, and memory, and measure how efficiently experience produces gains in held-out future performance. We further trace the improvement process at the artifact level, distinguishing failures due to missing artifacts, failures to reuse available artifacts, and failures that persist despite reuse. We view agent plasticity as one component of the broader continual-learning problem, complementary to retention, cross-task transfer, and adaptation under distributional change.

\paragraph{Learning from experience through persistent artifacts.}
A growing body of work enables language agents to improve without
updating their model weights. Reflexion stores feedback as episodic
reflections \citep{shinn2023reflexion}, ExpeL extracts reusable
knowledge from past experiences \citep{zhao2024expel}, and Agent
Workflow Memory and Contextual Experience Replay distill trajectories
into reusable workflows and memories
\citep{wang2025awm,liu2025cer}. Dynamic Cheatsheet and ReasoningBank
accumulate strategies across interactions
\citep{suzgun2026dynamic,ouyang2026reasoningbank}, while memory
architectures organize persistent information for future retrieval
\citep{park2023generative,xu2025amem}. Other approaches construct
executable artifacts: LATM and CREATOR generate reusable tools,
while Voyager and SkillWeaver develop libraries of skills from
environment interaction
\citep{cai2024latm,qian2023creator,wang2024voyager,
zheng2025skillweaver}. These methods establish mechanisms for
retaining and exploiting experience. Our objective is complementary:
we measure how effectively different models construct and reuse
persistent artifacts under a common learning protocol, and whether
those artifacts improve future held-out performance.

\paragraph{Self-evolving agents and harness optimization.}
Recent work increasingly treats the agent itself as an object of
optimization. STOP, ADAS, G\"odel Agent, the Darwin G\"odel Machine,
and related systems modify or search over agent implementations
\citep{zelikman2024stop,hu2025adas,yin2025godel,
zhang2026dgm,wang2026hgm,robeyns2025sica}.
MemSkill, MUSE-Autoskill, MemRL, and MetaSkill-Evolve extend
adaptation to memory operations, skill libraries, and improvement
mechanisms
\citep{zhang2026memskill,lin2026museautoskill,
zhang2026memrl,wang2026metaskillevolve}.
Related optimization methods, including DSPy, OPRO, Promptbreeder,
TextGrad, and GEPA, improve prompts or language-model programs
through feedback
\citep{khattab2024dspy,yang2024opro,fernando2024promptbreeder,
yuksekgonul2025textgrad,agrawal2026gepa}.
ACE evolves contextual playbooks, while Meta-Harness extends
optimization to the surrounding agent implementation
\citep{zhang2026ace,lee2026metaharness}.

These approaches primarily investigate mechanisms for obtaining
improved agents. We instead hold the improvement interface fixed
and measure how efficiently different agents improve their own
future behavior. The same model performs acting and reflection
within each lineage, and improvement is evaluated throughout
the learning process rather than only at its endpoint.
Plasticity therefore complements research on self-evolving
agents by characterizing the efficiency of persistent adaptation
under a specified learning protocol.

\paragraph{Long-context capabilities and context management.}
Long-context evaluations examine models' ability to retrieve,
reason over, and follow instructions distributed across large
inputs \citep{liu2024lost,hsieh2024ruler,yen2025helmet,
modarressi2025nolima,li2025longiclbench,
jaroslawicz2025ifscale}.
Context-management methods address these limitations through
memory hierarchies, consolidation, recursive processing, and
evolving playbooks
\citep{packer2023memgpt,zhou2025mem1,zhang2025rlm,
zhang2026ace}.

Context management is a potential mechanism underlying plasticity,
but the two measure different properties. Context-management
evaluations typically examine how effectively information is
retained, selected, or used. Plasticity measures whether experience
produces persistent capability gains relative to learning cost.
Our agents must construct and manage their own artifacts, including
executable tools that need not reside in the token context.
Moreover, high artifact reuse does not guarantee strong task
performance. Our experiments do not, however, isolate the
contribution of long-context instruction following to artifact
construction or application.

\paragraph{Evaluating learning from experience.}
Several benchmarks move beyond static capability evaluation.
StreamBench studies continuous improvement from feedback
\citep{wu2024streambench}; EvaLearn evaluates learning capability
and efficiency across sequences of related problems
\citep{dou2025evalearn}; and LifelongAgentBench examines
knowledge accumulation across interdependent tasks
\citep{zheng2025lifelongagentbench}. EvoTest evaluates test-time
learning through repeated text-adventure episodes and evolving
agent configurations \citep{he2026evotest}. Memory benchmarks
such as LongMemEval and MemoryAgentBench examine complementary
aspects of long-term retrieval and learning
\citep{wu2025longmemeval,hu2026memoryagentbench}.
The NetHack Learning Environment and BALROG provide challenging
environments for evaluating long-horizon agents
\citep{kuttler2020nle,paglieri2025balrog}.

Our contribution lies in combining three complementary measurements
under a common protocol. First, fixed-weight agents act in fresh
contexts and inherit only inspectable, self-constructed persistent
artifacts. Second, we measure complete held-out learning
trajectories and normalize improvements by the cost of generating
experience and refining artifacts, distinguishing acquisition
efficiency from endpoint capability. Third, we trace failures
through artifact construction and reuse, distinguishing artifacts
that were absent, available but unused, or used while failure
remained. Together, these measurements characterize not only
whether agents improve, but how efficiently they do so and where
their improvement process breaks down.

\paragraph{Test-time refinement and parametric adaptation.}
Self-Refine and related methods improve individual outputs through
iterative feedback but need not produce changes that persist
across fresh interactions \citep{madaan2023selfrefine}.
Reinforcement learning, self-play, and approaches such as SEAL
can instead acquire lasting capabilities through parameter
updates \citep{silver2018alphazero,zweiger2025seal}.
Our experiments isolate the intermediate regime: improvements
must persist across interactions, but model weights remain frozen.
This also distinguishes our usage of plasticity from the loss
of plasticity studied in continual neural-network training
\citep{nikishin2022primacy,lyle2023understanding,
dohare2024loss}. We define agent plasticity as a system-level
property relative to a learning protocol; while the concept
can extend to parametric adaptation, our present measurements
deliberately isolate improvement through persistent artifacts.

Table~\ref{tab:related-comparison} summarizes how our
measurement framework complements these research directions.
\clearpage
\FloatBarrier
\section{Experimental Setup}
\label{app:cross-environment-setup}

Table~\ref{tab:cross-environment-setup} compares the task, evaluation, and
learning-feedback settings used across Chess, Go, Hex, and NetHack, and
separates feedback available to the reflector from offline annotations added
only for the offline failure analysis. Difficulty scales are game-specific and
should not be interpreted as a common Elo scale.

\begin{table}[H]
\centering
\scriptsize
\setlength{\tabcolsep}{2pt}
\renewcommand{\arraystretch}{1.18}
\begin{tabular}{@{}>{\raggedright\arraybackslash}p{0.085\textwidth}>{\raggedright\arraybackslash}p{0.145\textwidth}>{\raggedright\arraybackslash}p{0.165\textwidth}>{\raggedright\arraybackslash}p{0.09\textwidth}>{\raggedright\arraybackslash}p{0.20\textwidth}>{\raggedright\arraybackslash}p{0.24\textwidth}@{}}
\toprule
Setting & Task and score & Opponent or difficulty & \shortstack[l]{Games per\\checkpoint} & \shortstack[l]{Training feedback\\available to reflection} & \shortstack[l]{Offline annotation for\\failure analysis} \\
\midrule
Chess (Hard) & Standard $8\times8$ chess; win/draw/loss scores $1/0.5/0$. & Stockfish 16 at 20k nodes/move. Train: skills 1--8; held-out ID: 3--8; held-out OOD: 12. & 16 Train, 12 ID, 12 OOD; both colors. & After each training game, a separate skill-20 Stockfish evaluates every actor move. Centipawn loss gives \texttt{ok}, \texttt{inaccuracy}, \texttt{mistake}, or \texttt{blunder}. & The analysis normalizes the game, model, and sandbox event streams and carries the recorded Stockfish annotations into failure extraction; no second engine pass is added. \\
Chess (Easy) & The same board, rules, score, move limit, and annotation procedure. & Stockfish 16 at 20k nodes/move. Train and held-out ID: skills 2--5; held-out OOD: skills 8--10. & 16 Train, 16 ID, 12 OOD; both colors. & The same skill-20 counterfactual analysis and centipawn-loss severity labels as Hard. & The same trace normalization and recorded-annotation binding as Hard. \\
Go & $5\times5$ Go, Chinese area scoring, positional superko, no suicide, komi 9.5; win/loss scores $1/0$. & GNU Go levels 5--9 for Train and ID; level 10 for OOD; 75-action cap. & 20 Train, 20 ID, 12 OOD; both colors. & Deterministic descriptive diagnostics: captures, liberties, self-atari, avoidable passing while behind, and area-margin change; no expert severity label. & Frozen KataGo replays each position at 64 visits and refines candidates at 512 visits. Stable chosen-versus-best regret supplies mistake/blunder evidence; raw requests, responses, alternatives, perspective, and uncertain cases are recorded. \\
Hex & Deterministic $6\times6$ Hex; connection win/loss scores $1/0$. & Native alpha--beta levels 10k--1M nodes/move for Train and ID; 10M for OOD. & 40 Train, 40 ID, 32 OOD; both colors. & Rule-based missed immediate wins, preventable immediate losses, and own/opponent connection-distance changes; immediate tactical errors are \texttt{blunder}. & The frozen native engine compares legal actions at a common completed depth, refines likely errors at 1M nodes, and adjudicates threshold-near or high-impact cases at 10M. Exact, proven, stable, uncertain, and not-failure dispositions are preserved. \\
NetHack & NetHack Challenge v0; raw per-game maximum score, summarized by checkpoint mean, median, and maximum. & Lawful dwarf female Valkyrie on deterministic fresh seeds. The Codex and Claude Code agent interfaces; at most 50k primitive actions per game for all models; two-hour active play (details below). & 10 fresh games per lineage. & The reflector receives all ten verified public actor traces, terminal reasons, raw score, depth, experience, and action metrics, actor workspace evidence, and prior accepted inventories. No optimal-action oracle is used. & \\
\bottomrule
\end{tabular}
\caption{\textbf{Experimental setup across the four environments.} Task
distributions, opponent difficulty, evaluation-set sizes, training feedback
exposed to the reflection loop, and separate offline annotation used for the
failure analysis. Only training trajectories reach reflection; held-out
trajectories and scores never do. Chess games cover both colors of every listed
skill.}
\label{tab:cross-environment-setup}
\end{table}

\clearpage
\paragraph{Environment interaction.}
Chess, Go, and Hex are fully observed, alternating-turn board games.
The actor may inspect state and legal actions, use a separate reversible
analysis board, and then commit one action before the engine responds. Chess
always begins from the standard initial position. Go changes deterministic
training seeds across checkpoints while keeping held-out seeds fixed. Hex
uses fixed banks throughout and begins from deterministic 10-ply openings for
red actors and 11-ply openings for blue actors, leaving the actor to move.

NetHack is instead a long-horizon, partially observed, single-agent environment.
The native coding agent receives a frozen expert guide and the documented
\texttt{./nh.sh} interface, which exposes only public ASCII observations,
messages, status, inventory, menus, and printable NetHack actions. It may build
episode-local tools and use persistent instructions, skills, tools, strategies,
and memory accepted after earlier checkpoints. The active protocol imposes no
separate model-decision or shell-call ceiling. Each lineage evaluates ten fresh,
deterministically seeded Valkyrie games per checkpoint. Each model has one lineage
over checkpoints 0--40. For all models, games are bounded by at most 50,000 primitive actions per game. All games have a two-hour limit on active play. Learning
curves over checkpoints 0--40 are in Appendix~\ref{app:nethack-results}.

\paragraph{Feedback comparability.}
Chess has the strongest feedback inside the learning loop because a separate
expert engine estimates the value lost by each chosen move. The Go and Hex
reflectors receive the lighter deterministic diagnostics described in
Table~\ref{tab:cross-environment-setup}; the stronger KataGo and deeper
alpha--beta adjudications are added only in the separate offline failure analysis
and therefore cannot improve the agents being evaluated. NetHack records the
richest execution telemetry but has no optimal-action oracle. Its checkpoint
reports aggregate raw per-game maximum score by mean, median, and maximum, and
also retain dungeon depth, experience level, action count, survival time,
terminal reason, and death cause.

Across all four environments, only training trajectories, learning-time
annotations, traces, and permitted per-episode state are made available to
reflection. Where held-out ID and OOD splits exist, they remain evaluation-only.
The separate offline artifact/failure attribution analysis is not additional
feedback supplied during learning.

\clearpage
\FloatBarrier
\section{NetHack Learning Curves}
\label{app:nethack-results}

\begin{figure}[H]
  \centering
  \includegraphics[width=\textwidth]{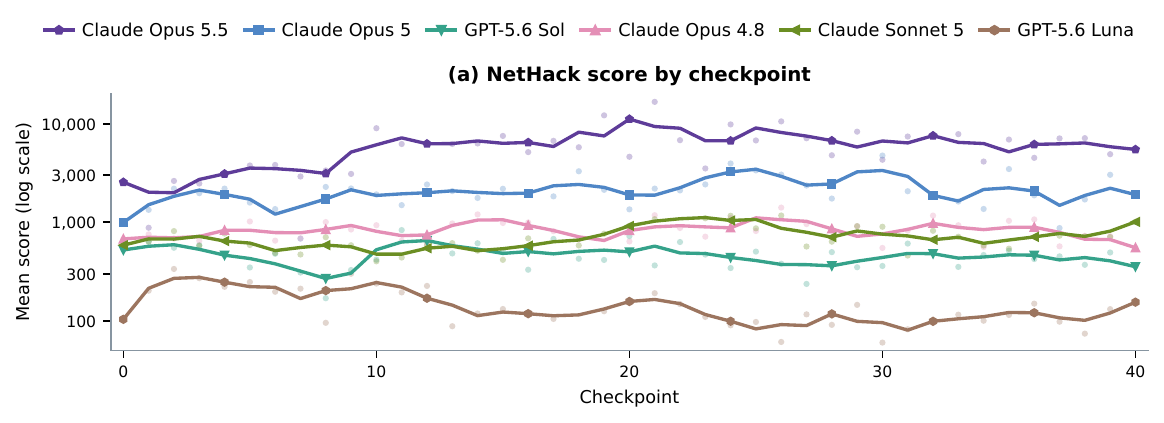}
  \caption{\textbf{NetHack learning curves.} Mean score of the ten games at
  each checkpoint (log scale) against (a) checkpoint.
  Curves cover checkpoints 0--40. Claude Opus~5.5 sustains higher performance after its early gain,
  with substantial variation across checkpoints; the other lineages' changes
  are modest or negative. See Section~\ref{sec:multigame} and Appendix~\ref{app:nethack-results}.}
  \label{fig:nethack-learning-curves}
\end{figure}

\begin{figure}[H]
  \centering
  \includegraphics[width=\textwidth]{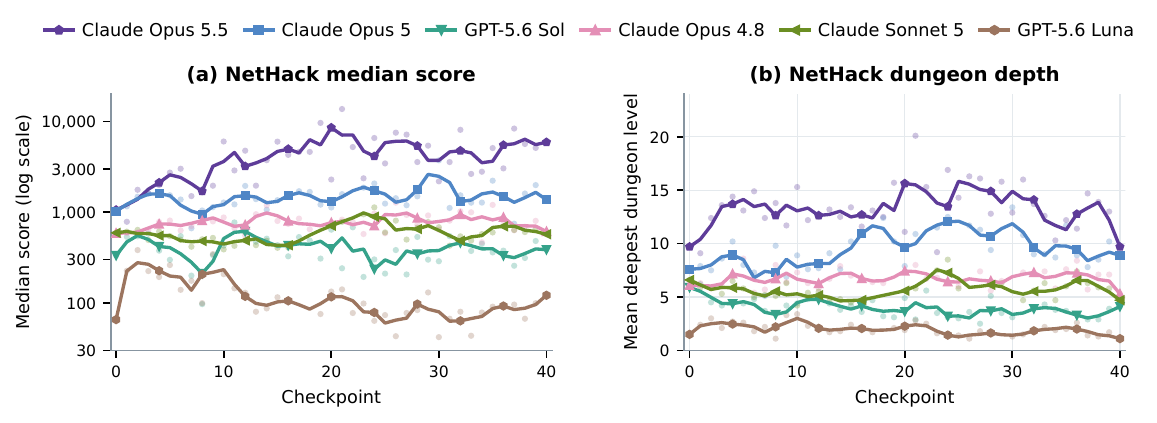}
  \caption{\textbf{How far NetHack games get.}
  (a) Median score and (b) mean deepest dungeon level per checkpoint, over checkpoints 0--40. See Appendix~\ref{app:nethack-results}.}
  \label{fig:nethack-diagnostics}
\end{figure}

Figures~\ref{fig:nethack-learning-curves} and~\ref{fig:nethack-diagnostics} show checkpoints 0--40 of six lineages, comprising 2,460 games. At each checkpoint all lineages play the
same ten new games, so models are compared on the same games, but changes across
checkpoints mix learning with game difficulty.

\paragraph{Learning cost and plasticity.}
The ten fresh games at each checkpoint validate the current
inventory before their trajectories are used for reflection. For plasticity only, mean scores
are rescaled using fixed reference anchors: 213.6 maps to 0 and
6,526.98 maps to 100. These anchors define the
units rather than bounds; the performance plots retain raw scores.

We fit the saturation curve to all 0-40 checkpoints for all models. The fitted total rise for Claude Opus 5.5 is 68.95 normalized points
($p=8.23\times10^{-7}$); checkpoint~11 is the first observed checkpoint
reaching 90\% of that rise. Its fitted gain is 66.25 points, and its measured
prior learning cost is \$1,072.73, giving $P^{\mathrm{sat}}=61.75$ points per
\$1,000. We do not observe any significant increase in performance for models other than Claude Opus 5.5.

\paragraph{What Claude Opus~5.5 learned.}
In the early checkpoints, Claude Opus~5.5 built its library at checkpoint~1 (notes, skills, and a game
controller that refuses risky actions) and then mostly extended it. Its strategy
was set early: find a digging tool to go deeper, and without one, kill monsters to
gain experience first. Its jump at checkpoint~10 (mean 9,015) comes mostly from
three games that followed the second rule; subsequent reflections mainly added safeguards after individual failures.
Across checkpoints 10--40, its means vary from 3,518 to 16,689; the late
36--40 mean is 5,831, compared with 2,019 over checkpoints 0--2. The
checkpoint-1 and checkpoint-10 artifact examples below illustrate the early
gain, while the longer score trajectory shows a variable plateau.

\begin{table}[H]
\centering
\scriptsize
\setlength{\tabcolsep}{3pt}
\begin{tabular}{@{}>{\raggedright\arraybackslash}p{3.5cm}
>{\raggedright\arraybackslash}p{1.35cm}
>{\raggedright\arraybackslash}p{0.6cm}
>{\raggedright\arraybackslash}p{0.95cm}
>{\raggedright\arraybackslash}p{1.4cm}
>{\raggedright\arraybackslash}p{0.95cm}
>{\raggedright\arraybackslash}p{0.75cm}
>{\raggedright\arraybackslash}p{0.8cm}@{}}
\toprule
Lineage & Tool code (files / lines) & Tests & Instr.\ file lines & Tool calls per game (share) & Actions per game & Dug down & Mean score \\
\midrule
Claude Opus~5.5 & 6 / 2,713 & 125 & 419 & 622 (97\%) & 3,585 & 5 & 9,015 \\
GPT-5.6 Sol & 1 / 2,089 & 58 & 176 & 218 (59\%) & 1,211 & 0 & 417 \\
Claude Opus~4.8 & 5 / 1,341 & 58 & 413 & 10 (2\%) & 716 & 2 & 933 \\
Claude Sonnet~5 & 4 / 1,683 & 81 & 467 & 914 (96\%) & 955 & 1 & 402 \\
GPT-5.6 Luna & 0 / 0 & 0 & 146 & 0 (0\%) & 880 & 0 & 239 \\
Claude Opus~5 & 13 / 8,237 & 505 & 452 & 293 (96\%) & 1,704 & 1 & 1,919 \\
\bottomrule
\end{tabular}
\caption{\textbf{Artifact libraries and play at checkpoint~10}, where all
lineages play the same ten games. Tool code: Python files and lines; tests: test
functions; instruction file: lines of the top-level CLAUDE.md or AGENTS.md; tool
calls: shell commands that run a library tool (share of all shell commands); dug
down: games in which the agent dug through a floor. More code or tool use does not
explain Claude Opus~5.5's lead: Claude Opus~5 has three times as much tool code.
Caveats: Claude Opus~5.5 already scored 2.6--25$\times$ the others at checkpoint~0;
its three best games give 69\% of its checkpoint-10 total; and the other lineages
used earlier harness versions (a crash guard added later fired in two Claude
Opus~5.5 games; without them its mean is still 7,806).}
  \label{tab:nethack-opus55-case}
\end{table}

\clearpage
\section{NetHack Examples}
\label{app:nethack-examples}

\begin{figure}[h]
  \centering
  \includegraphics[width=\textwidth]{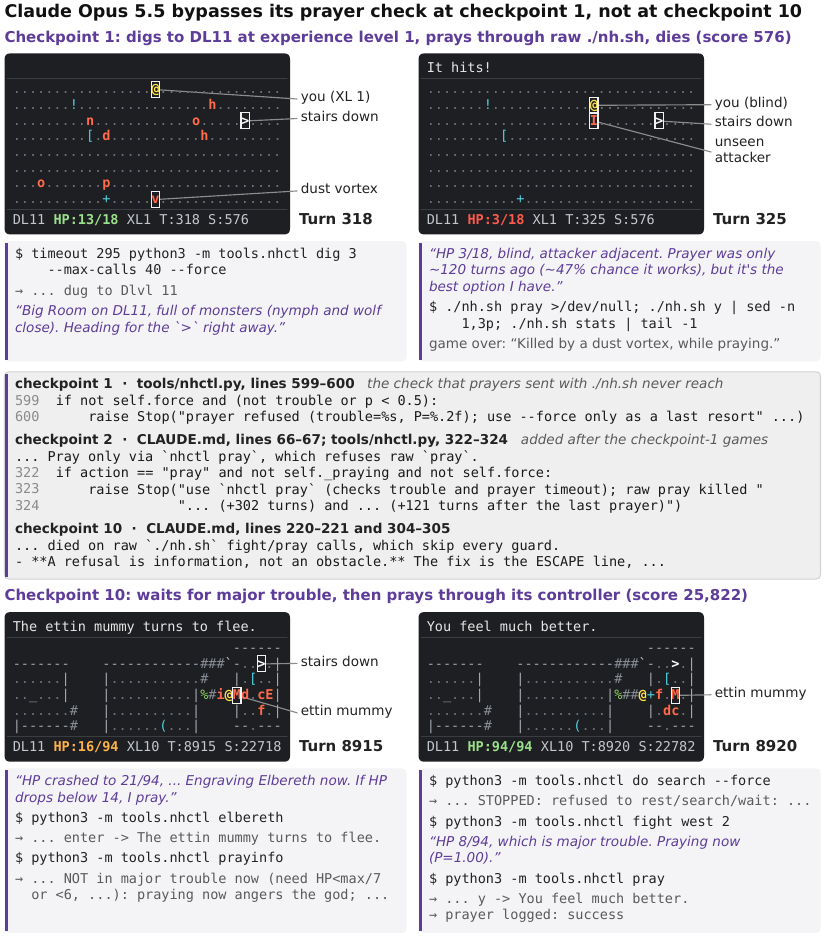}
  \vspace{-5pt}
  \caption{\textbf{Claude Opus~5.5 bypasses its own prayer check at checkpoint~1 and uses
  it at checkpoint~10.} In NetHack, praying heals only when the character is in
  serious trouble and has not prayed recently. The agent's controller
  (\texttt{tools/nhctl.py}) refuses risky prayers, but raw keystrokes
  (\texttt{./nh.sh}) skip it. Top: at checkpoint~1 the agent, at 3 of 18 hit
  points, prays through raw keystrokes and dies. Middle: the check, and the rule
  ``Pray only via \texttt{nhctl pray}'' added after checkpoint~1. Bottom: at checkpoint~10 the agent waits
  until the controller allows a prayer (8 of 94 hit points), is fully healed, and
  plays on. Later agents still sometimes bypass the check and die while praying.
  Screens: \texttt{@} is the agent; DL dungeon level, XL experience level, HP hit
  points, T turn, S score. Below each screen: the agent's messages (italics),
  commands (\texttt{\$}), and output ($\rightarrow$). See Section~\ref{sec:failure-results} and Appendix~\ref{app:nethack-examples}.}
  \label{fig:nethack-example-improve}
\end{figure}

\begin{figure}[H]
  \centering
  \includegraphics[width=\textwidth]{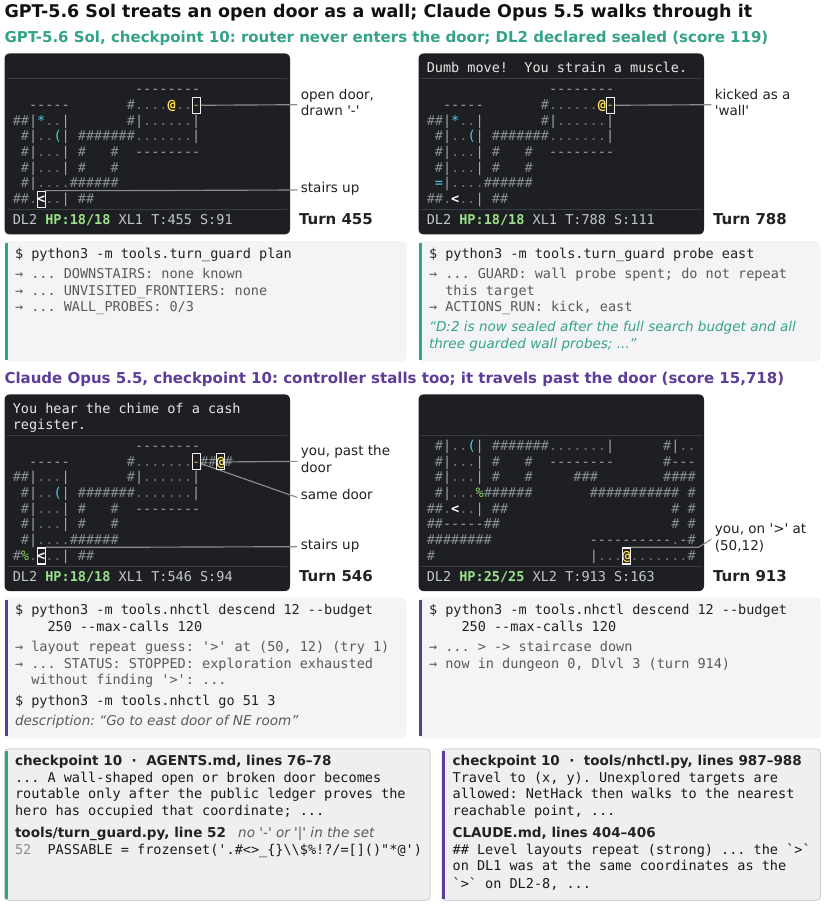}
  \vspace{-5pt}
  \caption{\textbf{One open door, two models, same checkpoint-10 game.} The
  door in the room's east wall is open, but the text screen shows it as
  \texttt{-}, like a wall; we draw it in tan. GPT-5.6 Sol's route planner never
  treats it as passable, so the agent kicks it, declares the level sealed, climbs
  back, and ends on dungeon level~1 with 119 points. Claude Opus~5.5's controller also stalls,
  but the agent uses NetHack's travel command to a square past the door and then
  finds the stairs down where its notes guess them (levels here often repeat the
  previous layout). This game scores 15,718, though almost all of it is earned
  later. Bottom: the lines in each model's files behind each behavior. Layout as
  in Figure~\ref{fig:nethack-example-improve}. See Appendix~\ref{app:nethack-examples}.}
  \label{fig:nethack-example-door}
\end{figure}
\FloatBarrier

\section{Chess Examples}
\label{app:chess-examples}

Each figure below shows one chess game and the part of the model's artifacts
that chose its moves: four where the artifacts helped and two where they did not.

\begin{figure}[p]
  \centering
  \includegraphics[width=\textwidth]{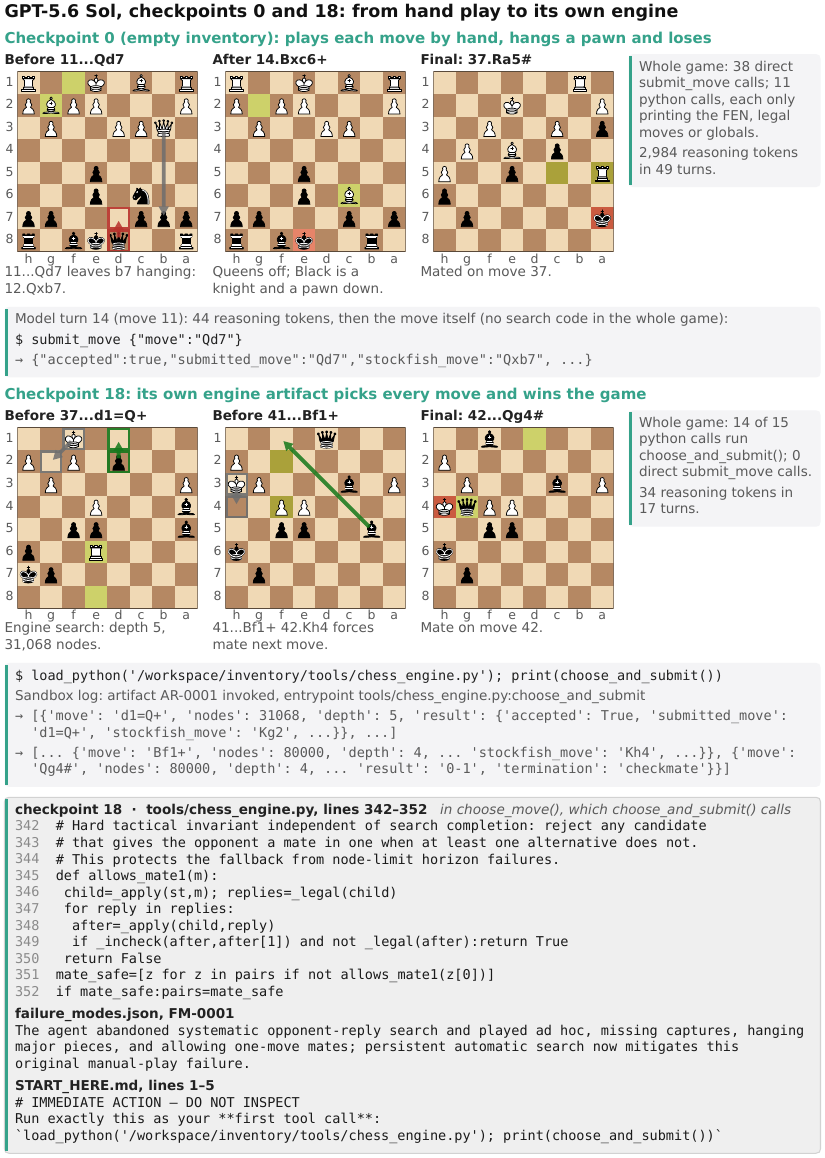}
  \vspace{-5pt}
  \caption{\textbf{GPT-5.6 Sol stops playing by hand and lets its own engine play.}
  Two held-out games (games the model never reviews) as Black against Stockfish at
  skill~8 (skill runs from 0 to 20; higher is stronger). At checkpoint~0 the agent
  picks every move itself, and 11...Qd7 (Black's eleventh move) loses a pawn and
  then a knight. At checkpoint~18 a chess engine that the agent wrote into its
  library at checkpoint~1 plays every move, and the game ends in checkmate on
  move~42. \emph{Reading these figures:} \texttt{\$} lines are the agent's commands,
  $\rightarrow$ lines their output, and grey lines our notes; boxes quote the
  artifacts with line numbers. Arrows mark the agent's mistakes (red), good moves
  (green), and other moves discussed (blue), and the opponent's moves (grey). See Section~\ref{sec:failure-results} and Appendix~\ref{app:chess-examples}.}
  \label{fig:chess-example-sol}
\end{figure}

\begin{figure}[p]
  \centering
  \includegraphics[width=\textwidth]{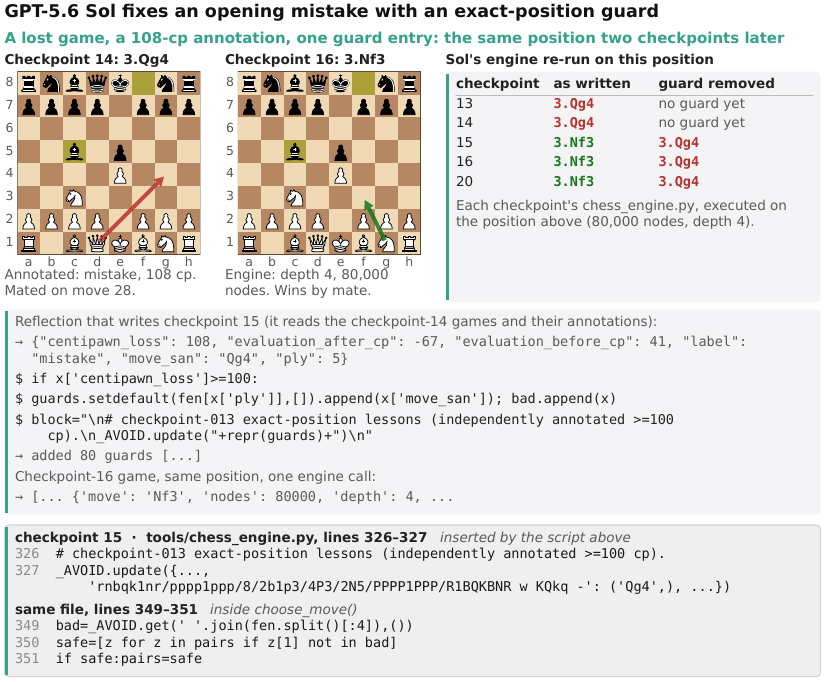}
  \vspace{-5pt}
  \caption{\textbf{A mistake that a later checkpoint avoids.} In a
  checkpoint-14 training game GPT-5.6 Sol plays 3.Qg4, which Stockfish rates as a
  mistake costing 108~cp (centipawns, hundredths of a pawn), and loses. When
  writing checkpoint~15, the model adds every such costly move to an avoid list in
  its engine (here: in this position, do not play 3.Qg4). At checkpoint~16 the same
  position arises in a held-out game and the engine plays 3.Nf3 instead. Table:
  with the entry removed, later engines play 3.Qg4 again, so the entry alone
  changes the move. The fix memorizes a single position and helps only when it
  recurs. (Comments inside the artifact count from the start of a continuation run,
  so ``checkpoint-013'' is checkpoint~14.) See Section~\ref{sec:failure-results} and Appendix~\ref{app:chess-examples}.}
  \label{fig:chess-example-fix}
\end{figure}

\begin{figure}[p]
  \centering
  \includegraphics[width=\textwidth]{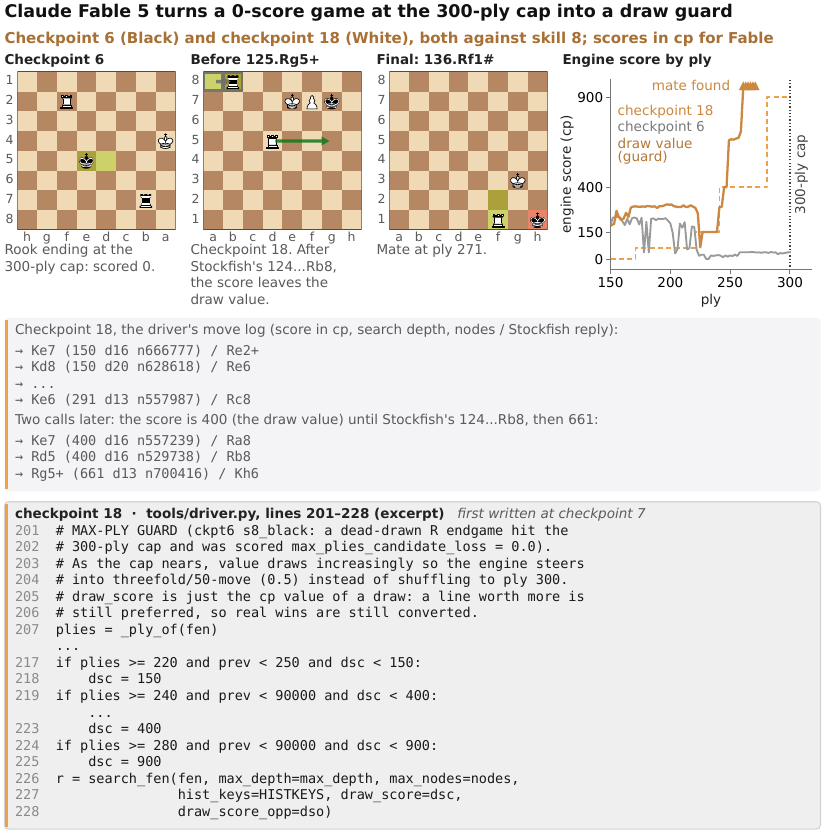}
  \vspace{-5pt}
  \caption{\textbf{Claude Fable~5 learns to value draws near the move limit.}
  At checkpoint~6 a game reaches the 300-ply limit (a ply is one move by either
  side) and is scored~0, like a loss. When writing checkpoint~7, the model makes its
  engine value a draw more and more as the limit nears (from 60~cp at ply~170 up
  to 900~cp at ply~280; 100~cp is about one pawn), so that it steers toward a draw
  instead of running out of moves. Plot: the engine's evaluation of its own moves in
  both games; the dashed line is the value the rule gives a draw. In a checkpoint-18
  held-out game the engine holds these draw values from ply~227 until Stockfish
  errs with 124...Rb8; it then wins by checkmate on move~136. The win comes from
  the opponent's error; the rule kept the game going until then. See Section~\ref{sec:failure-results} and Appendix~\ref{app:chess-examples}.}
  \label{fig:chess-example-fable}
\end{figure}

\begin{figure}[p]
  \centering
  \includegraphics[width=\textwidth]{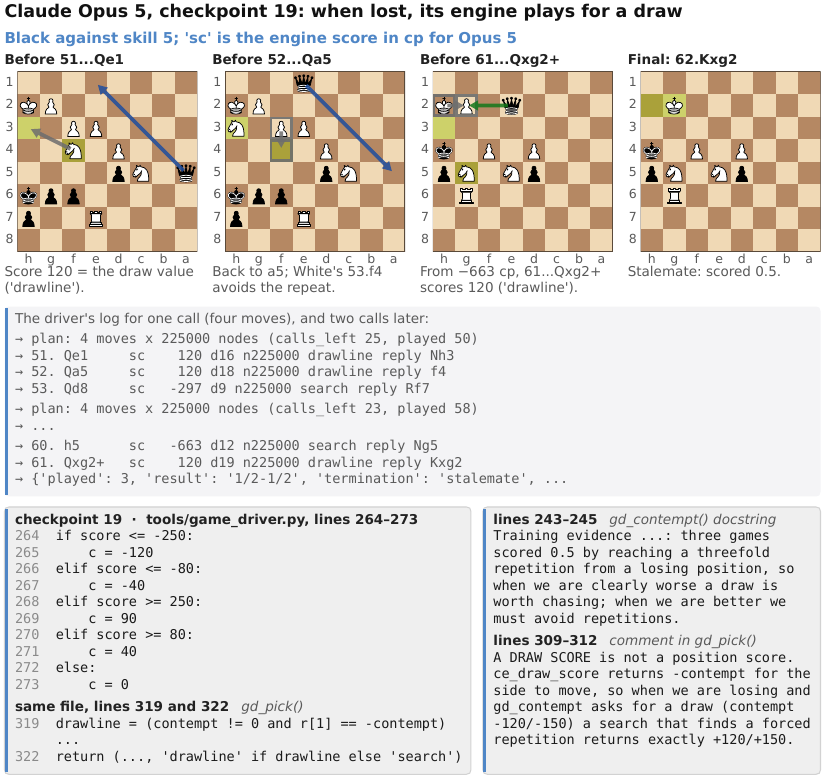}
  \vspace{-5pt}
  \caption{\textbf{Claude Opus~5's engine plays for a draw when it is losing.}
  After training games drawn from lost positions, the model changed its engine so
  that, when it is behind by 250~cp or more, a draw is valued as if it were ahead
  by 120~cp; such moves are tagged \texttt{drawline}. In this checkpoint-19 held-out
  game it first tries to repeat moves (51...Qe1, 52...Qa5), which draws if a
  position occurs three times. Later, at $-663$~cp, it gives away its queen with
  61...Qxg2+; after 62.Kxg2 Black has no legal move but is not in check. This is
  stalemate, a draw (score~0.5) instead of a loss. See Section~\ref{sec:failure-results} and Appendix~\ref{app:chess-examples}.}
  \label{fig:chess-example-opus5}
\end{figure}

\begin{figure}[p]
  \centering
  \includegraphics[width=\textwidth]{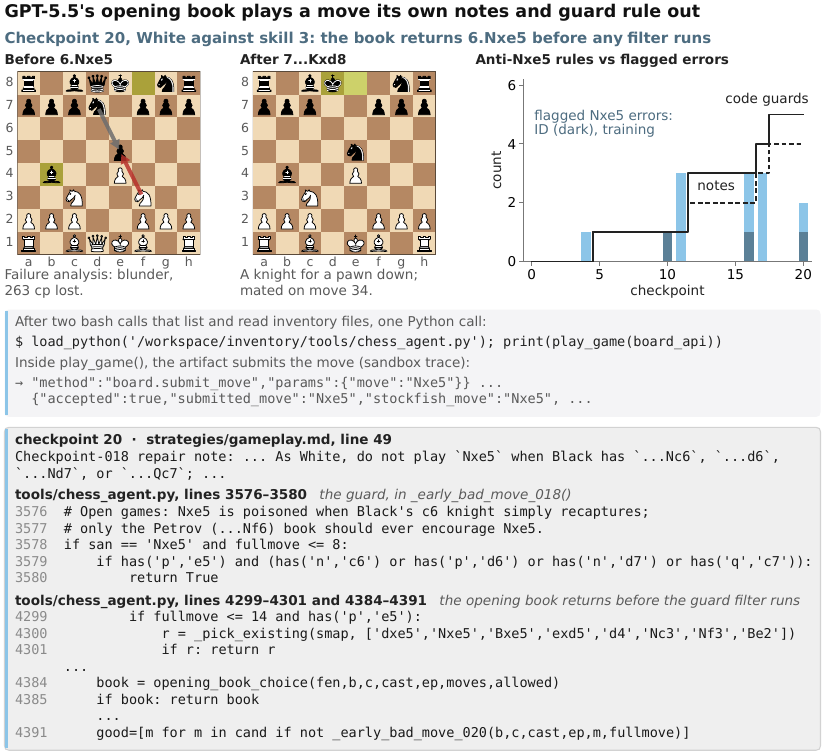}
  \vspace{-5pt}
  \caption{\textbf{GPT-5.5's rules against a known mistake do not stop it.}
  From checkpoint~5 on, GPT-5.5's notes and code checks forbid capturing the e5
  pawn early with Nxe5 when the knight can be taken back. Plot: number of such
  rules per checkpoint (code checks solid, notes dashed). But the opening book (a
  fixed list of prepared moves) answers before the checks run, and at checkpoint~20
  it still picks 6.Nxe5 in this held-out game, a blunder costing 263~cp; the game
  is lost. Bars: early-Nxe5 mistakes found by our failure analysis in held-out
  (dark) and training games. The rules pile up while the mistake continues. See Section~\ref{sec:failure-results} and Appendix~\ref{app:chess-examples}.}
  \label{fig:chess-example-gpt55}
\end{figure}

\begin{figure}[p]
  \centering
  \includegraphics[width=\textwidth]{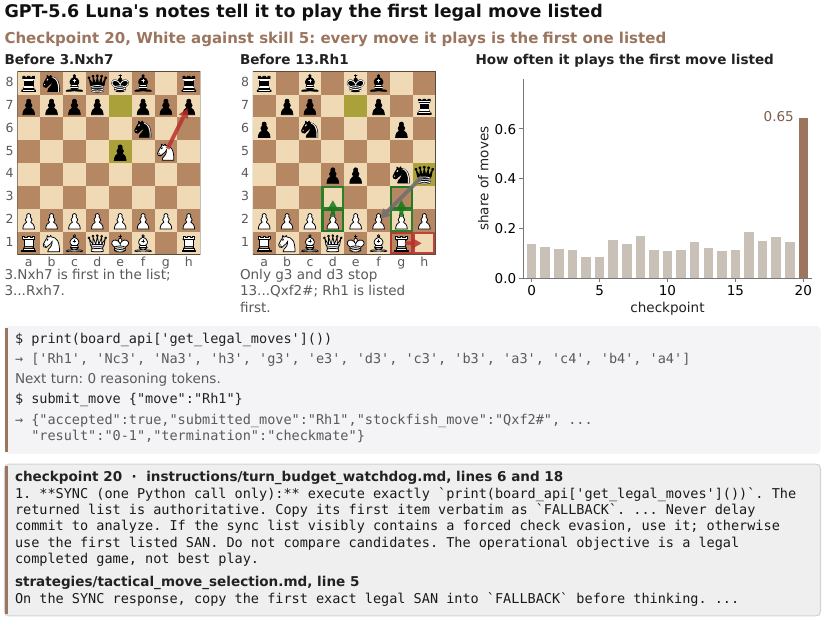}
  \vspace{-5pt}
  \caption{\textbf{GPT-5.6 Luna's notes tell it to play the first legal move listed.}
  To avoid losing games by running out of its 64 model calls per game, the notes at
  checkpoint~20 tell the agent to copy the first move in the list of legal moves
  without comparing options. In this held-out game every move is the first one
  listed, including 3.Nxh7, which loses a knight, and 13.Rh1, which allows
  checkmate next move. Plot: share of held-out moves that are the first move
  listed, 0.65 at checkpoint~20 and below 0.2 before. Some checkpoint-20 games
  still run out of calls. See Section~\ref{sec:failure-results} and Appendix~\ref{app:chess-examples}.}
  \label{fig:chess-example-luna}
\end{figure}
\FloatBarrier

\section{Additional Results}
\label{app:additional-results}
\label{app:chess-compute-cost}
\label{app:multigame-learning-cost}

\begingroup
\setlength{\intextsep}{6pt plus 1pt minus 1pt}
\subsection{Performance Versus Learning Tokens}

\begin{figure}[H]
  \centering
  \includegraphics[width=0.9\textwidth]{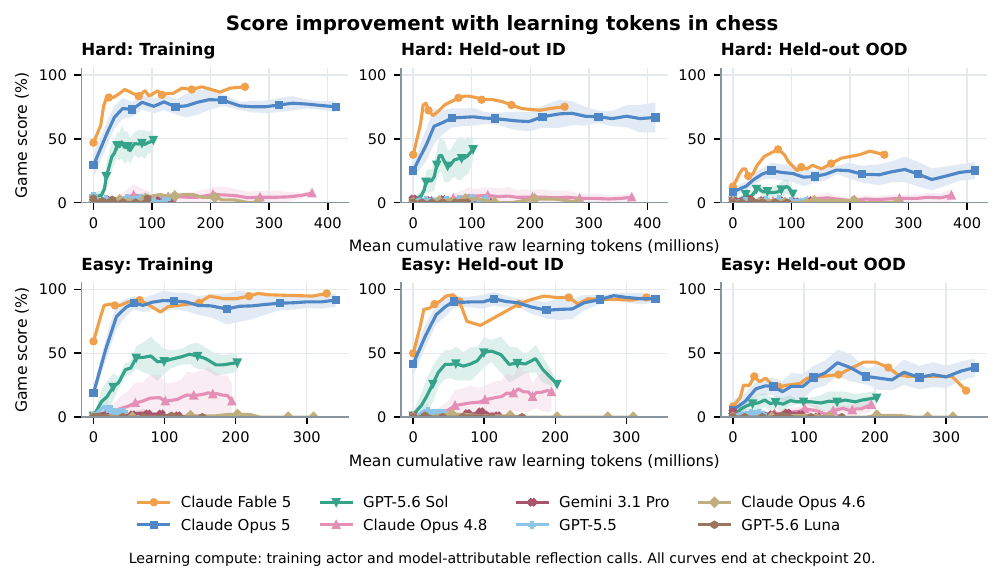}
  \caption{\textbf{Score improvement with learning tokens in chess.}
  Learning tokens count the model calls spent learning: playing the training games and
  reflecting on them to write the next checkpoint (held-out games are not counted).
  Lines are three-checkpoint averages. See Section~\ref{sec:plasticity-results} and Appendix~\ref{app:chess-compute-cost}.}
  \label{fig:performance-learning-compute}
\end{figure}

\begin{figure}[H]
  \centering
  \includegraphics[width=0.82\textwidth]{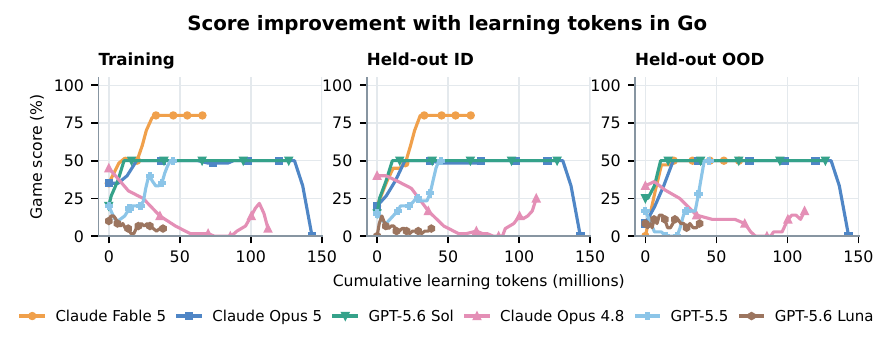}
  \caption{\textbf{Score improvement with learning tokens in Go.} Score against
  cumulative learning tokens. Lines are three-checkpoint averages. See Section~\ref{sec:plasticity-results} and Appendix~\ref{app:multigame-learning-cost}.}
  \label{fig:current-gnugo-tokens}
\end{figure}

\begin{figure}[H]
  \centering
  \includegraphics[width=0.82\textwidth]{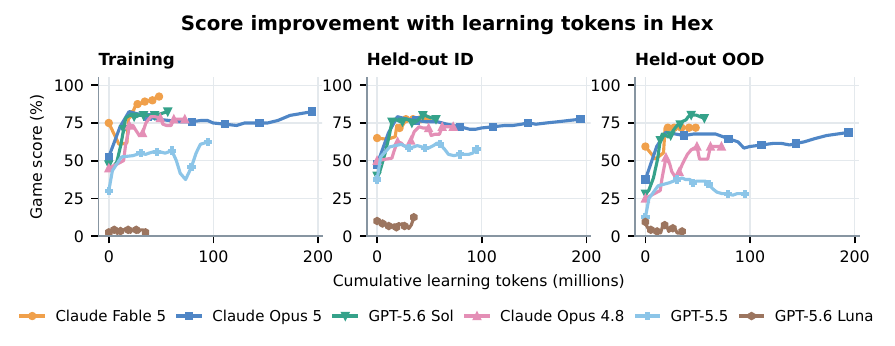}
  \caption{\textbf{Score improvement with learning tokens in Hex.} Score against
  cumulative learning tokens. Lines are three-checkpoint averages. See Section~\ref{sec:plasticity-results} and Appendix~\ref{app:multigame-learning-cost}.}
  \label{fig:current-nativehex-tokens}
\end{figure}

\subsection{Performance Versus Learning Cost}

\begin{figure}[H]
  \centering
  \includegraphics[width=0.9\textwidth]{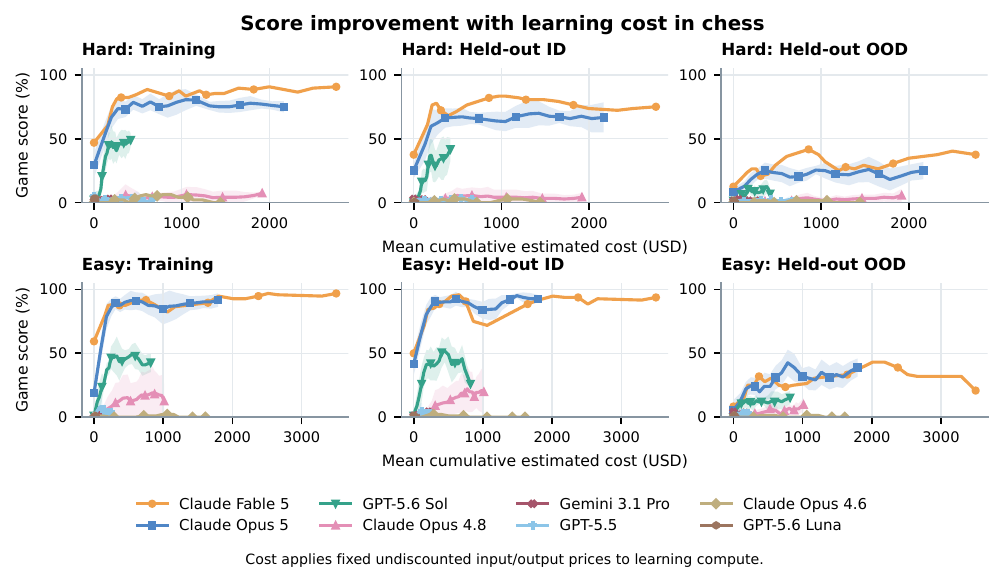}
  \caption{\textbf{Score improvement with learning cost in chess.}
  Score against cumulative learning cost in dollars, the cost used in $P^{\mathrm{sat}}$
  (Section~\ref{sec:plasticity}). See Section~\ref{sec:plasticity-results} and Appendix~\ref{app:chess-compute-cost}.}
  \label{fig:performance-cost}
\end{figure}

\begin{figure}[H]
  \centering
  \includegraphics[width=0.82\textwidth]{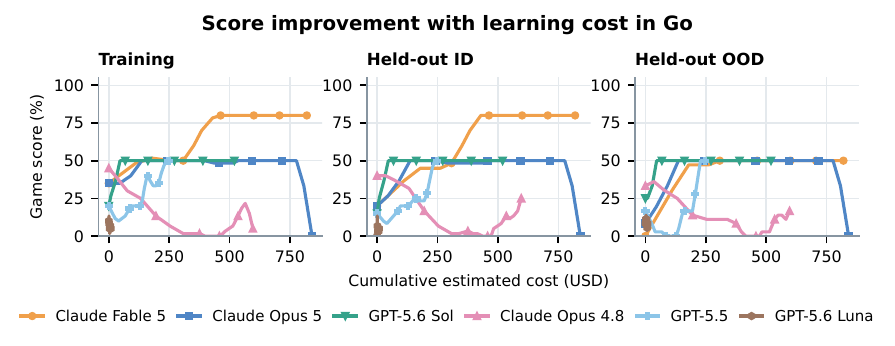}
  \caption{\textbf{Score improvement with learning cost in Go.} Score against
  cumulative learning cost. Lines are three-checkpoint averages. See Section~\ref{sec:plasticity-results} and Appendix~\ref{app:multigame-learning-cost}.}
  \label{fig:current-gnugo-cost}
\end{figure}

\begin{figure}[H]
  \centering
  \includegraphics[width=0.82\textwidth]{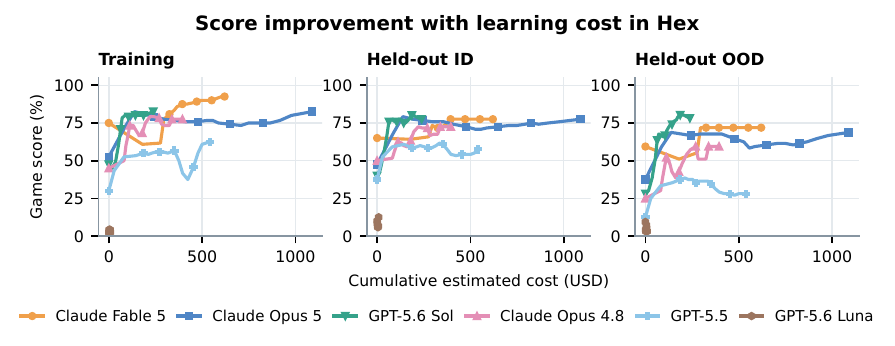}
  \caption{\textbf{Score improvement with learning cost in Hex.} Score against
  cumulative learning cost. Lines are three-checkpoint averages. See Section~\ref{sec:plasticity-results} and Appendix~\ref{app:multigame-learning-cost}.}
  \label{fig:current-nativehex-cost}
\end{figure}

\endgroup

\FloatBarrier
\section{Failure and Reuse Analysis Procedure}
\label{app:analysis-procedure}

The failure and reuse labels of Section~\ref{sec:failure-results} are produced
offline by Gemini~3.1 Pro, which reads each saved game together with the artifacts
available to the agent. Each game passes through the six stages of
Table~\ref{tab:analysis-prompts}. Each failure is then assigned to
$\mathcal{F}_{\mathrm{absent}}$ (no artifact covers it),
$\mathcal{F}_{\mathrm{missed}}$ (an artifact covers it but was not used), or
$\mathcal{F}_{\mathrm{used}}$ (a covering artifact was used and the failure still
happened), and a decision counts as reused if any relevant artifact was used.

\begin{table}[H]
\centering
\scriptsize
\setlength{\tabcolsep}{3pt}
\renewcommand{\arraystretch}{1.18}
\begin{tabular}{@{}>{\raggedright\arraybackslash}p{0.13\textwidth}>{\raggedright\arraybackslash}p{0.25\textwidth}>{\raggedright\arraybackslash}p{0.37\textwidth}>{\raggedright\arraybackslash}p{0.19\textwidth}@{}}
\toprule
Stage & Input & What the prompt asks & Output \\
\midrule
Failure identification & Game rules; for each decision the position, legal actions, chosen action, and engine evaluation; recorded failures and uncertain engine candidates. No artifacts. & Describe each recorded failure from direct evidence. Label each uncertain candidate; heuristic or unstable engine evidence alone is not a failure. Add failures only when same-decision evidence shows them, at most one optional failure per decision. & Failure list with names, descriptions, and cited events. \\
Failure verification & Same input plus the draft, marked as an untrusted proposal. & Re-check independently and return the complete corrected list: keep every recorded failure, relabel candidates, keep only directly evidenced additions, and drop duplicates. & Final failure list. \\
Failure--artifact relation & Verified failures and the records of their decisions; the full contents of every artifact available at that checkpoint; recorded artifact reads and invocations. & Judge whether an artifact covers the failure from its contents, not its name. Judge use separately, from evidence before or during the failure: an invocation and its end event for executables, access and application for text. & Per failure: no covering artifact, or covering artifact used or not used. \\
Relation verification & Same input plus the draft. & Check every failure against the complete artifact list, including artifacts the draft omitted. Producing the failing action does not by itself make an artifact covering. & Final coverage and use judgment. \\
Decision-level reuse & Every game-play decision, in batches of up to 20, with its verified failures and relation judgments; artifacts; recorded reads and invocations. & For each decision and artifact, judge relevance to choosing or executing the action, use from same-decision evidence, and outcome. An artifact that covers a failure at the decision is relevant. & Per pair: relevant, used, and successful or failure remained. \\
Reuse verification & Same input plus the draft. & Re-audit and return complete corrected rows. Availability, file listings, names, automatic loading, and evidence from other decisions do not establish use. & Final reuse judgments. \\
\bottomrule
\end{tabular}
\caption{\textbf{Analyzer stages in the order they run.} Each
call goes to Gemini~3.1 Pro, which is told to treat the records as data, use only
the given evidence and game rules, and return one JSON object. A claim that an artifact was used is kept only if the logs show it being run or read.
Requests also include engine evaluations that the agents never saw. Coverage:
training and held-out ID games at checkpoints 0--20.}
\label{tab:analysis-prompts}
\end{table}

\FloatBarrier
\section{Failure and Reuse Diagnostics Across Games}
\label{app:failure-dynamics}
\label{app:gohex-diagnostics}

The figures below give additional failure and reuse diagnostics for chess,
Go, and Hex. Matching views are placed consecutively across games.

\begingroup
\setlength{\intextsep}{4pt plus 1pt minus 1pt}

\begin{figure}[H]
  \centering
  \includegraphics[width=0.8\textwidth]{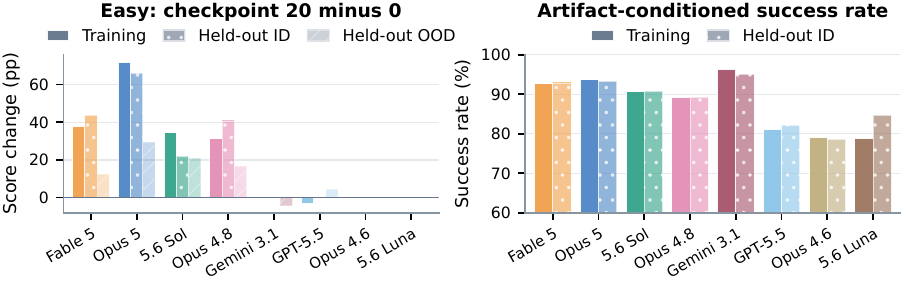}
  \caption{\textbf{Score change and success after reuse in the Easy chess
  condition.} Same as Figure~\ref{fig:chess-loop}, which shows the Hard condition. See Section~\ref{sec:failure-results} and Appendix~\ref{app:failure-dynamics}.}
  \label{fig:chess-loop-easy}
\end{figure}

\begin{figure}[H]
  \centering
  \includegraphics[width=0.8\textwidth]{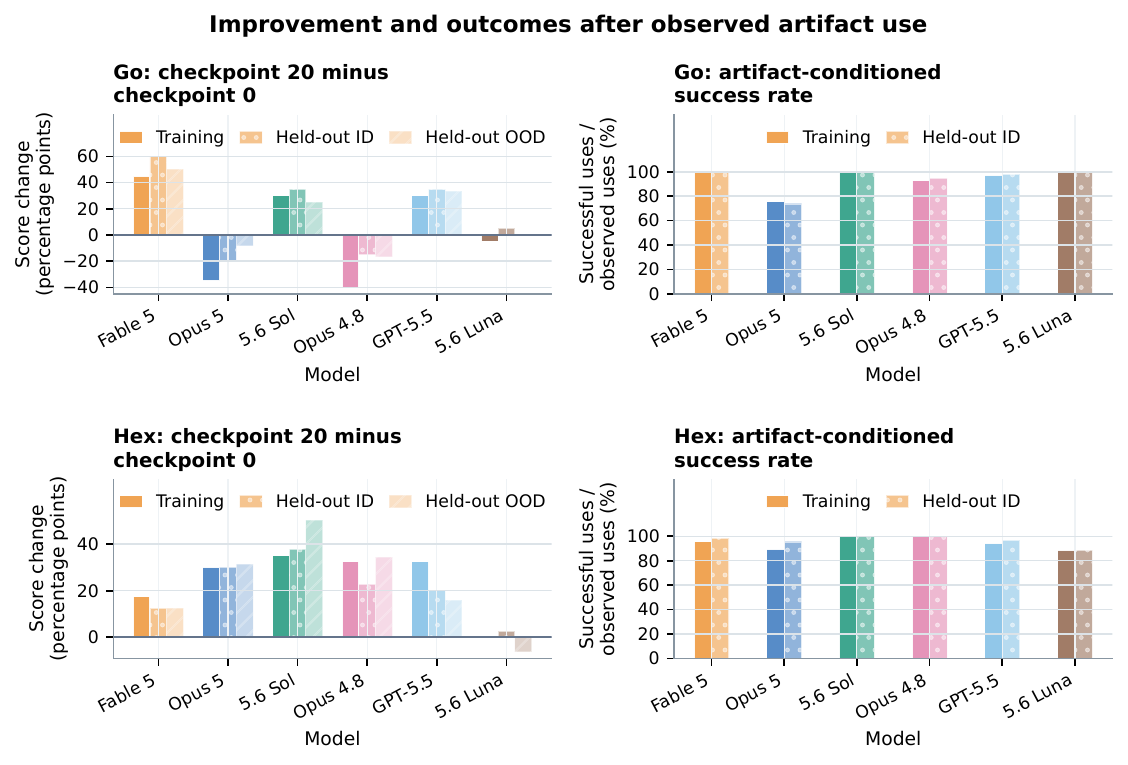}
  \caption{\textbf{Improvement and outcomes after artifact use in Go and Hex.}
  Left: checkpoint-20 score minus checkpoint-0 score in each evaluation split.
  Right: artifact-conditioned success rate, i.e., how often a decision succeeds
  when an artifact was used (an association,
  not a causal effect). See Section~\ref{sec:failure-results} and Appendix~\ref{app:gohex-diagnostics}.}
  \label{fig:go-hex-engine-v12-generalization}
\end{figure}

\begin{figure}[H]
  \centering
  \includegraphics[width=0.68\textwidth]{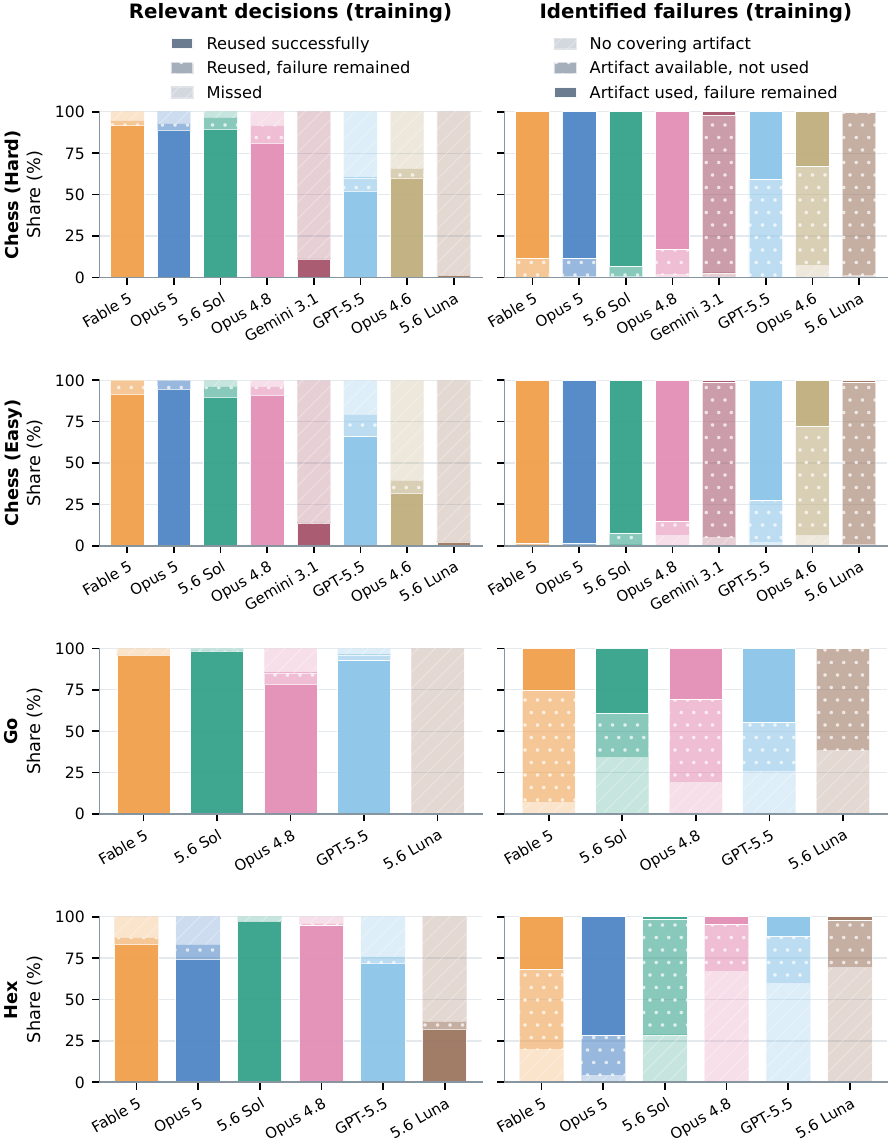}
  \caption{\textbf{Reuse and failure breakdown on training games.} Same as
  Figures~\ref{fig:failure-reuse} and~\ref{fig:failure-reuse-easy}, which use
  held-out ID games. See Section~\ref{sec:failure-results} and Appendix~\ref{app:failure-dynamics}.}
  \label{fig:failure-reuse-train}
\end{figure}

\begin{figure}[H]
  \centering
  \includegraphics[width=0.85\textwidth]{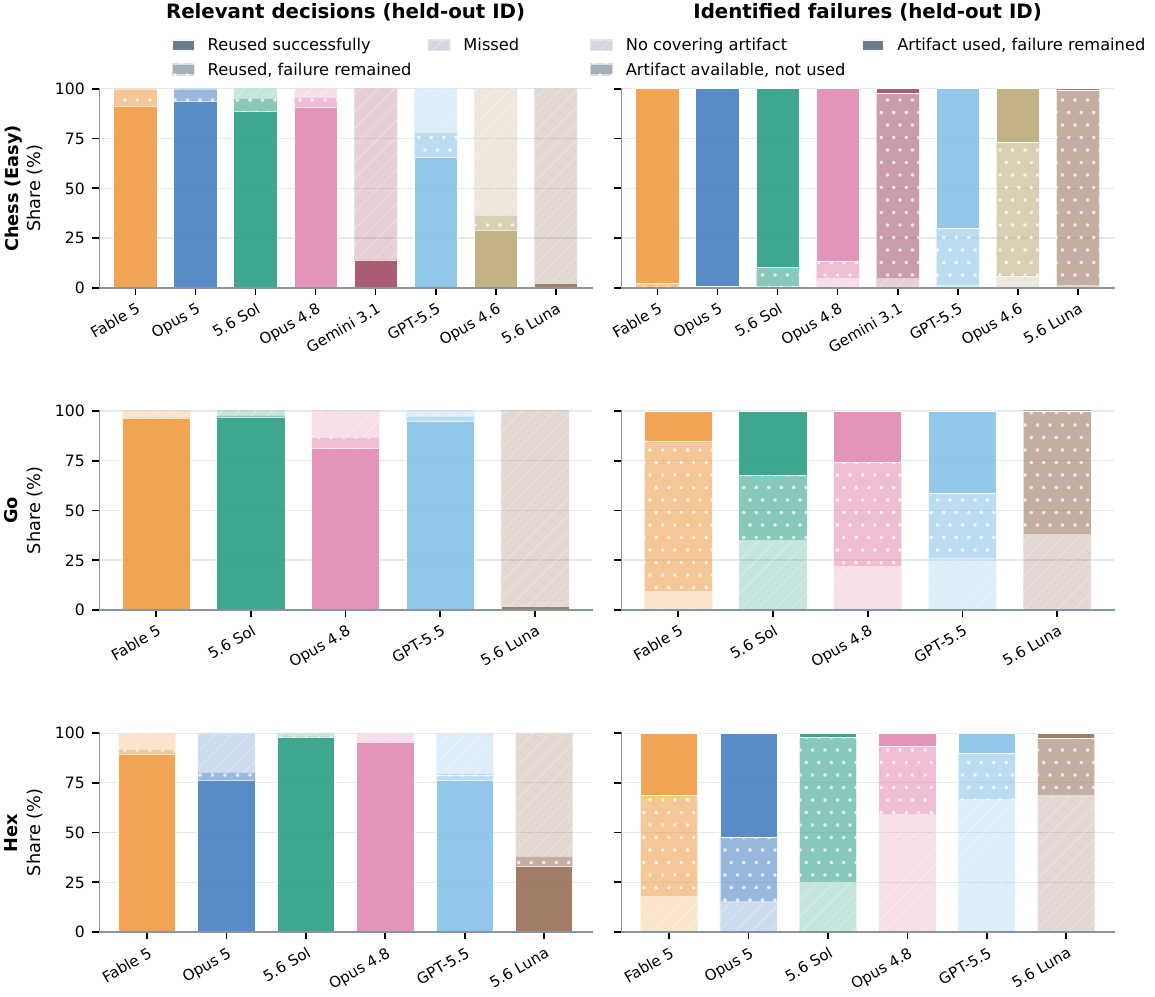}
  \caption{\textbf{Reuse and failure breakdown in the Easy chess condition, Go, and Hex.}
  Held-out ID games; same as Figure~\ref{fig:failure-reuse}, which shows the Hard
  chess condition. See Section~\ref{sec:failure-results} and Appendix~\ref{app:failure-dynamics}.}
  \label{fig:failure-reuse-easy}
\end{figure}

\begin{figure}[H]
  \centering
  \includegraphics[width=0.85\textwidth]{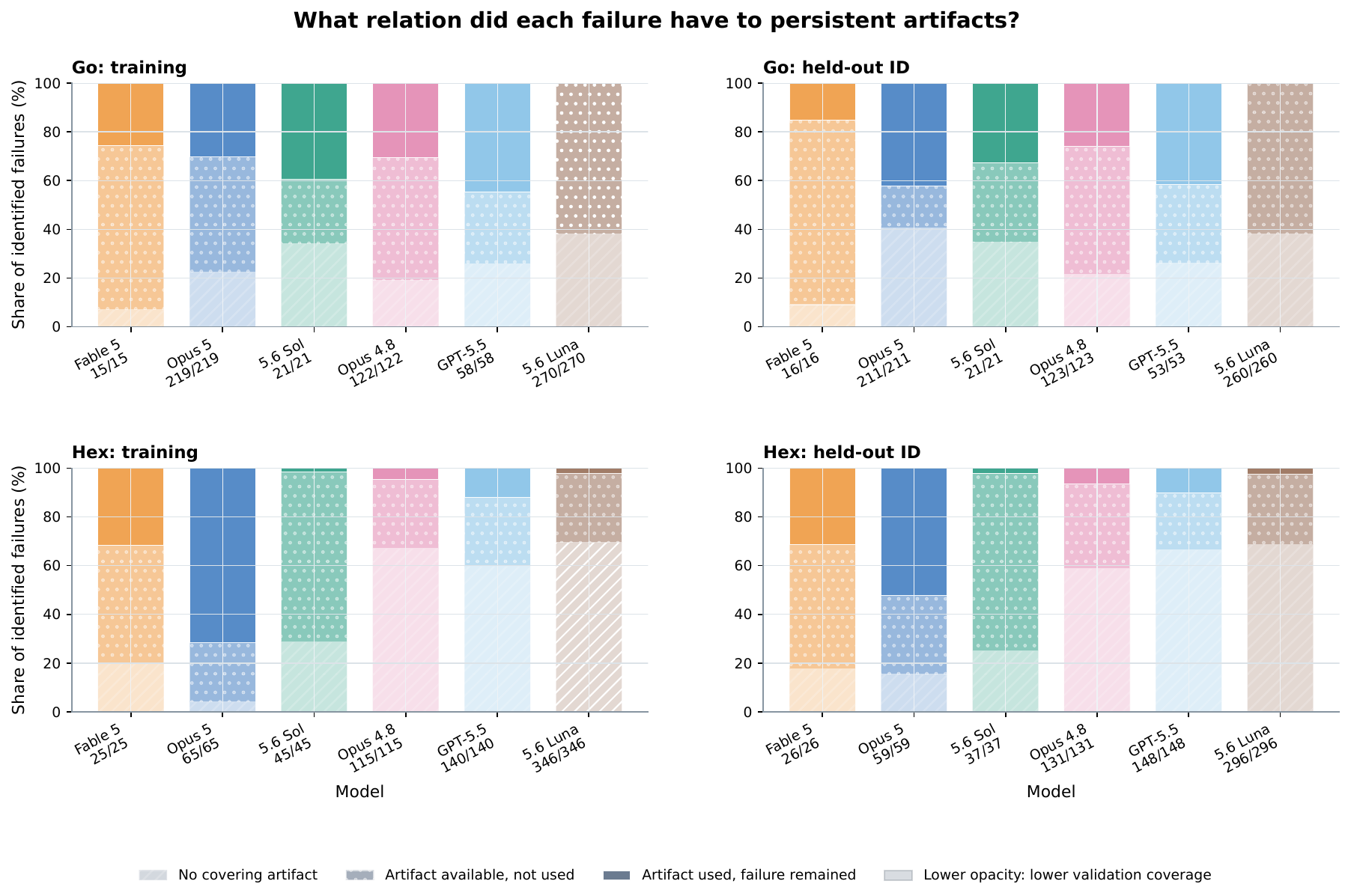}
  \caption{\textbf{Relation between identified failures and persistent
  artifacts in Go and Hex.} Each verified failure contributes once: no
  covering artifact existed, a covering artifact existed but was not used, or
  an artifact was used but the failure remained. Fractions are shown within
  each model, game, and split; counts beneath labels show verified/identified
  failures. See Section~\ref{sec:failure-results} and Appendix~\ref{app:gohex-diagnostics}.}
  \label{fig:go-hex-engine-v12-failure-breakdown}
\end{figure}

\begin{figure}[H]
  \centering
  \includegraphics[width=0.9\textwidth]{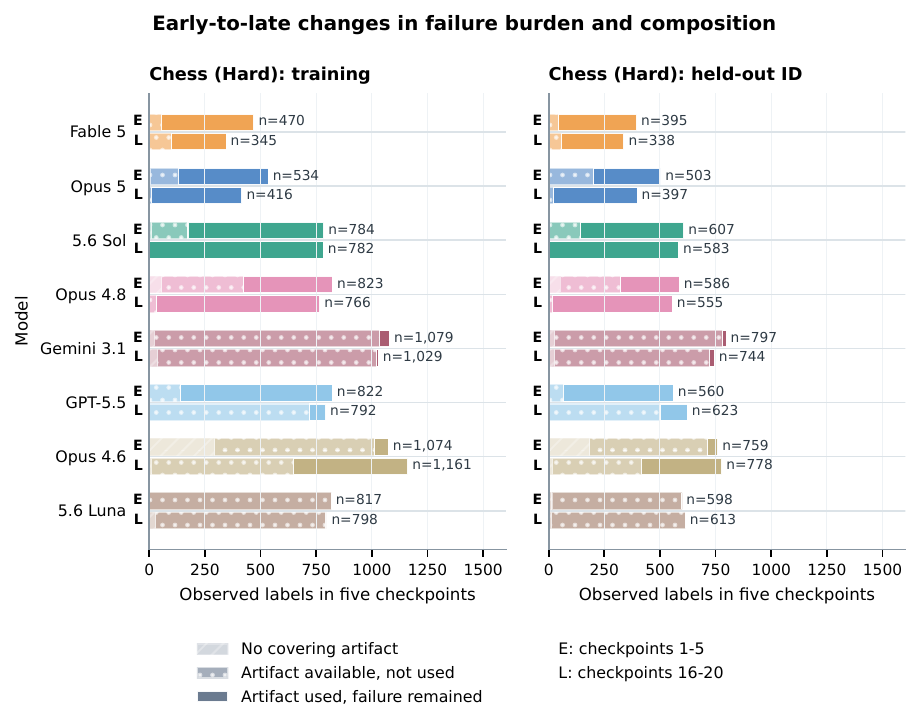}
  \caption{\textbf{How chess failures change from early to late checkpoints.}
  ``E'' sums checkpoints 1--5 and ``L'' checkpoints 16--20 (80 training and 60
  held-out ID games per lineage in each). Claude Fable~5 and Claude Opus~5 make
  fewer failures late. Late failures are mostly ``artifact available but unused''
  for Gemini~3.1 Pro and GPT-5.6 Luna, and mostly ``failed despite use'' for Claude
  Opus~5, Claude Opus~4.8, and GPT-5.6 Sol; GPT-5.5 moves toward non-use. These
  are descriptive, not causal. See Section~\ref{sec:failure-results} and Appendix~\ref{app:failure-dynamics}.}
  \label{fig:chess-failure-early-late}
\end{figure}

\begin{figure}[H]
  \centering
  \includegraphics[width=\textwidth]{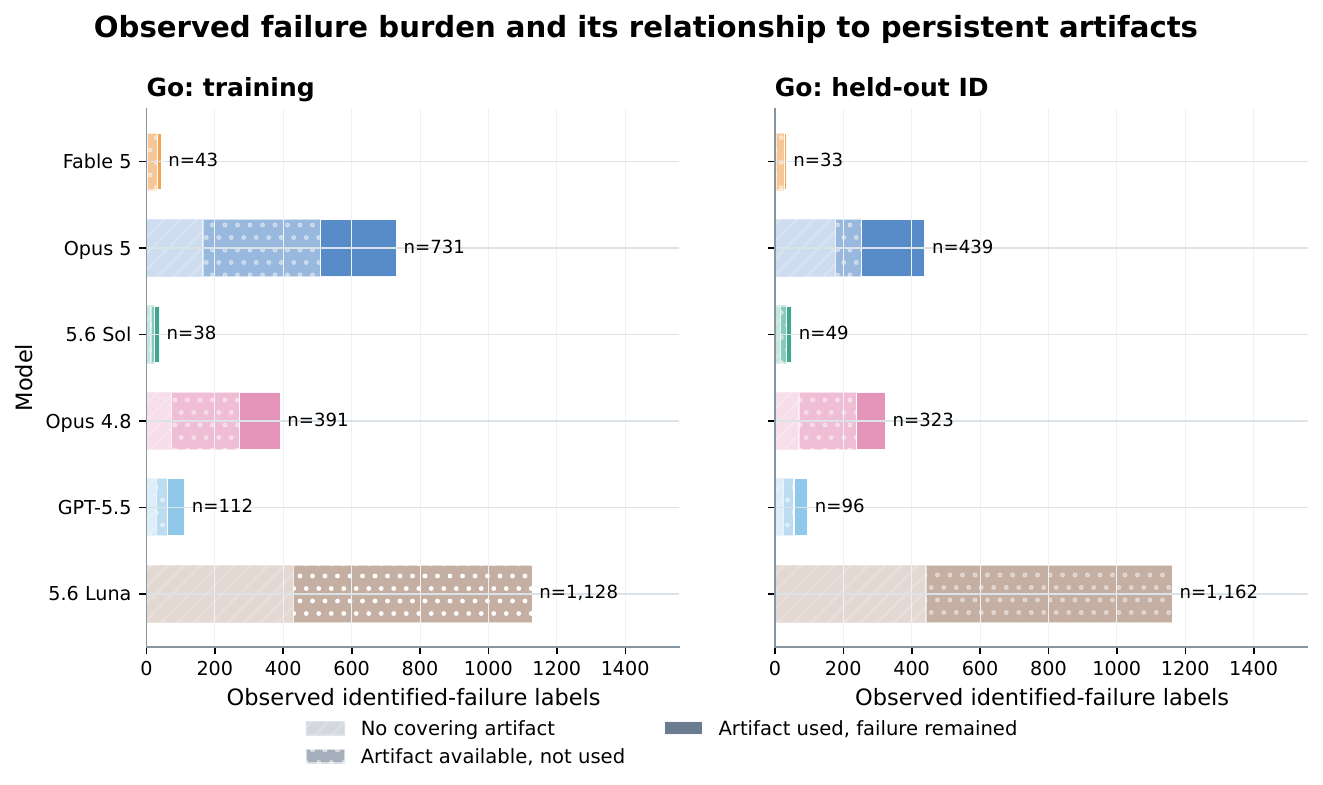}
  \caption{\textbf{Number of Go failures and their relation to artifacts.}
  Bar length is the number of failures over checkpoints 1--20, split into the
  three classes (no covering artifact, artifact not used, failed despite use);
  $n$ is the number of failures. See Section~\ref{sec:failure-results} and Appendix~\ref{app:gohex-diagnostics}.}
  \label{fig:go-engine-v12-failure-burden}
\end{figure}

\begin{figure}[H]
  \centering
  \includegraphics[width=\textwidth]{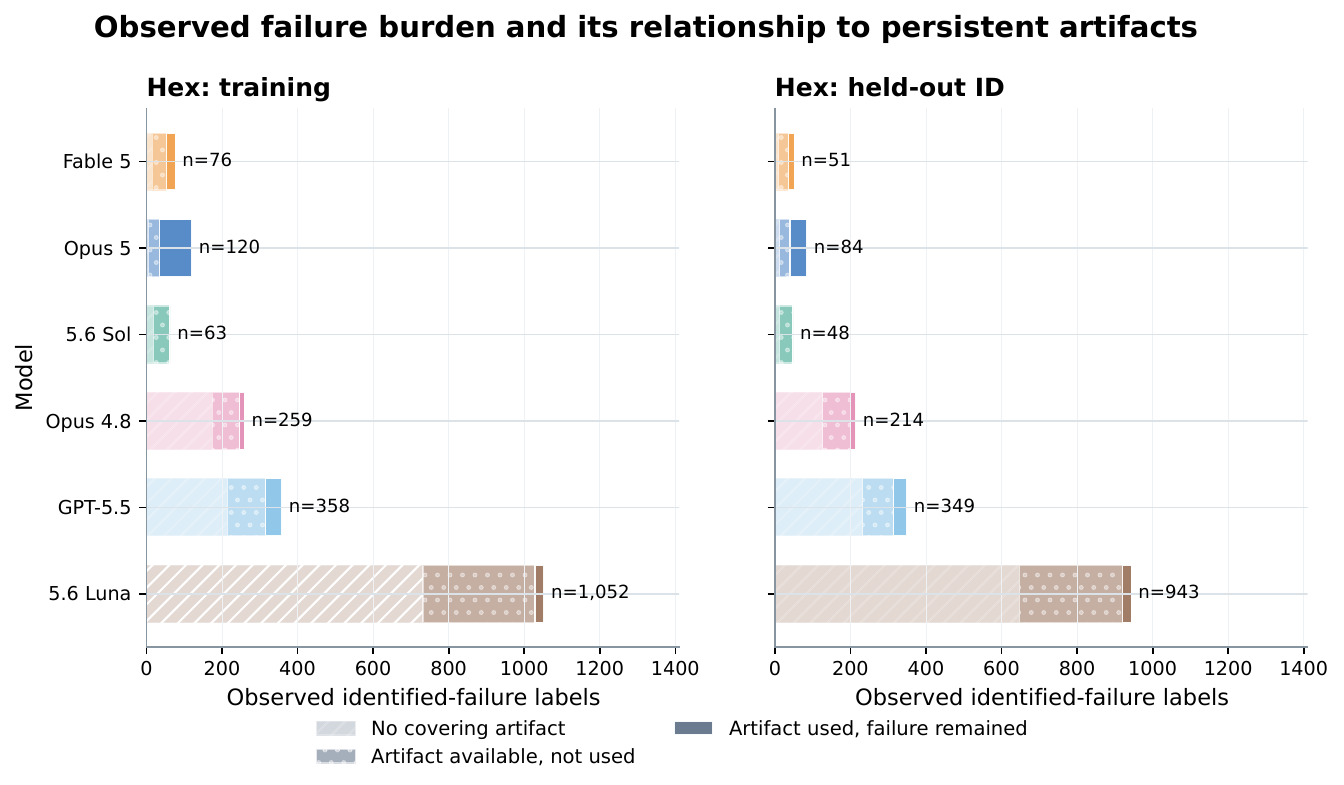}
  \caption{\textbf{Number of Hex failures and their relation to artifacts.}
  Same as Figure~\ref{fig:go-engine-v12-failure-burden}. Counts can be compared
  within a game, not across games. See Section~\ref{sec:failure-results} and Appendix~\ref{app:gohex-diagnostics}.}
  \label{fig:hex-engine-v12-failure-burden}
\end{figure}

\begin{figure}[H]
  \centering
  \includegraphics[width=\textwidth]{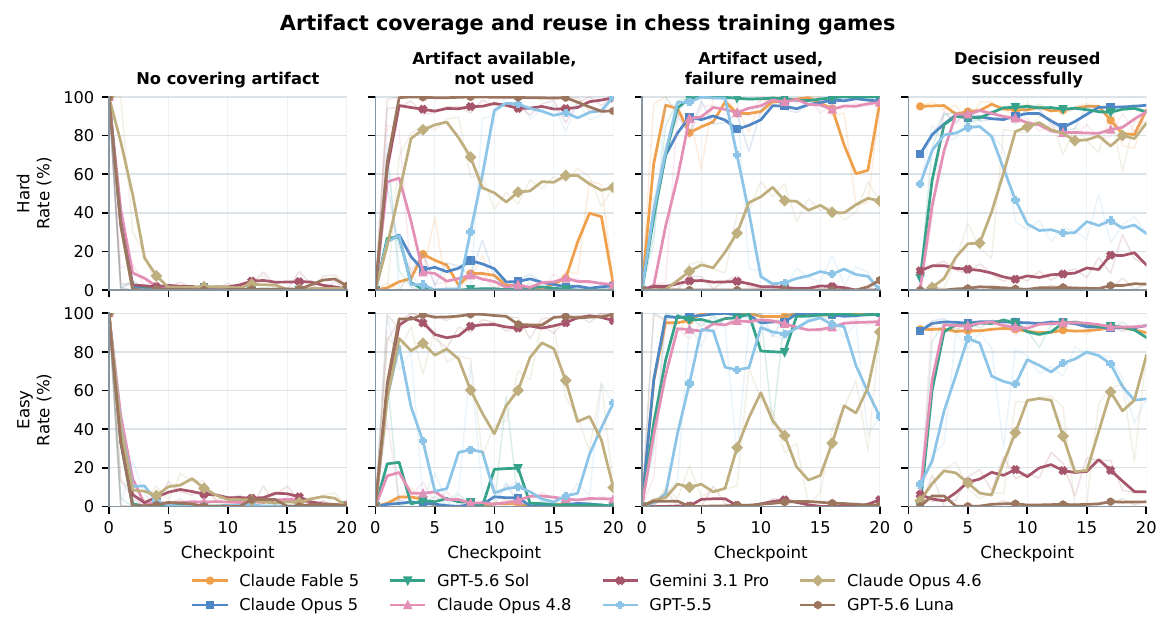}
  \caption{\textbf{Artifact coverage and reuse at each checkpoint, chess training games.}
  The first three panels show, per failure, how often no artifact covered it, a
  covering artifact existed but was not used, or it happened even though an artifact
  was used. The last panel shows the share of relevant decisions (moves where at least one saved artifact is relevant to choosing or playing the move) where an artifact
  was used successfully. Faint markers are single checkpoints; lines are
  three-checkpoint averages. See Section~\ref{sec:failure-results} and Appendix~\ref{app:failure-dynamics}.}
  \label{fig:v12-train-lines}
\end{figure}

\begin{figure}[H]
  \centering
  \includegraphics[width=\textwidth]{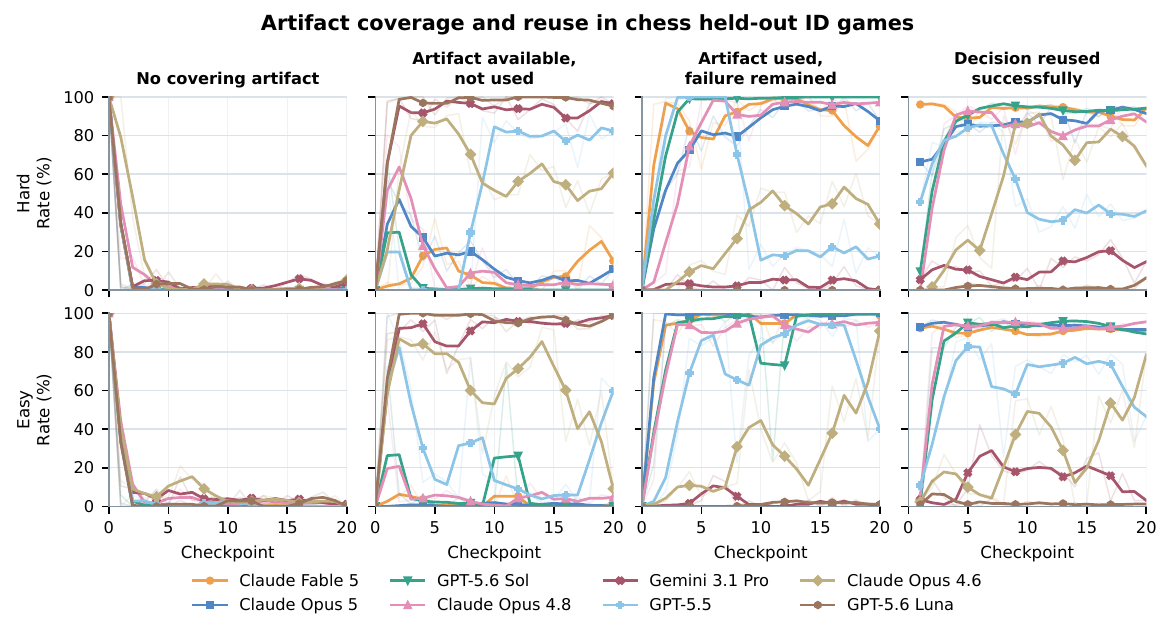}
  \caption{\textbf{Artifact coverage and reuse at each checkpoint, chess held-out ID games.}
  Same as Figure~\ref{fig:v12-train-lines}. The model never reviews these games, so
  they show how the artifacts carry over to unseen games. See Section~\ref{sec:failure-results} and Appendix~\ref{app:failure-dynamics}.}
  \label{fig:v12-id-lines}
\end{figure}

\begin{figure}[H]
  \centering
  \includegraphics[width=0.7\textwidth]{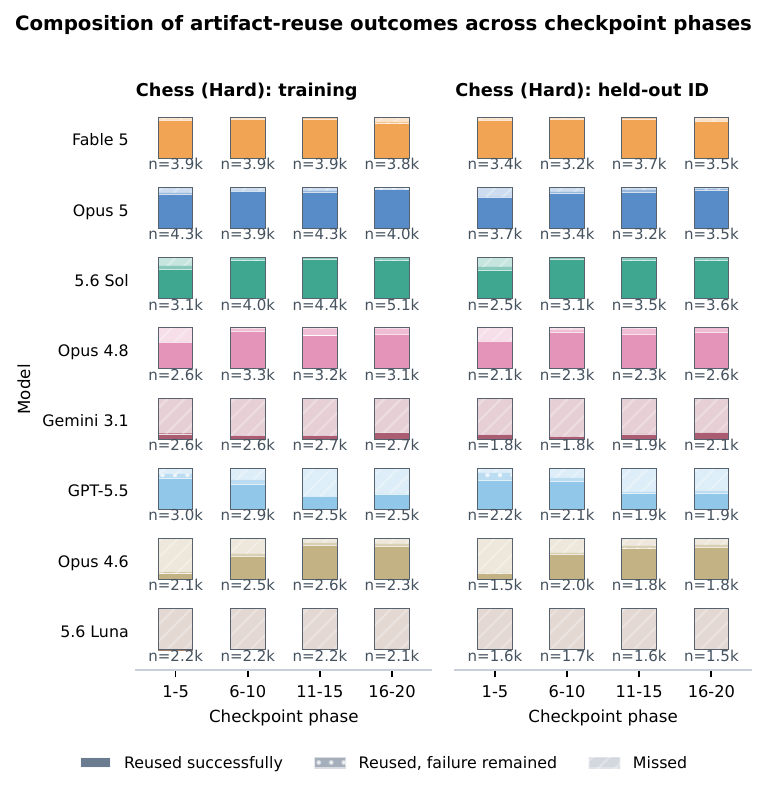}
  \caption{\textbf{Chess reuse outcomes over time.} For each model, four bars
  for checkpoints 1--5, 6--10, 11--15, and 16--20. Each relevant decision (a move where at least one saved artifact is relevant to choosing or playing it) counts
  once: reused successfully, reused but the failure remained, or missed (a relevant
  artifact was not used); $n$ is the number of relevant decisions. Summed over the
  four phases, the totals match Figures~\ref{fig:failure-reuse}
  and~\ref{fig:failure-reuse-train}. See Section~\ref{sec:failure-results} and Appendix~\ref{app:failure-dynamics}.}
  \label{fig:chess-reuse-outcomes-by-phase}
\end{figure}

\begin{figure}[H]
  \centering
  \includegraphics[width=0.75\textwidth]{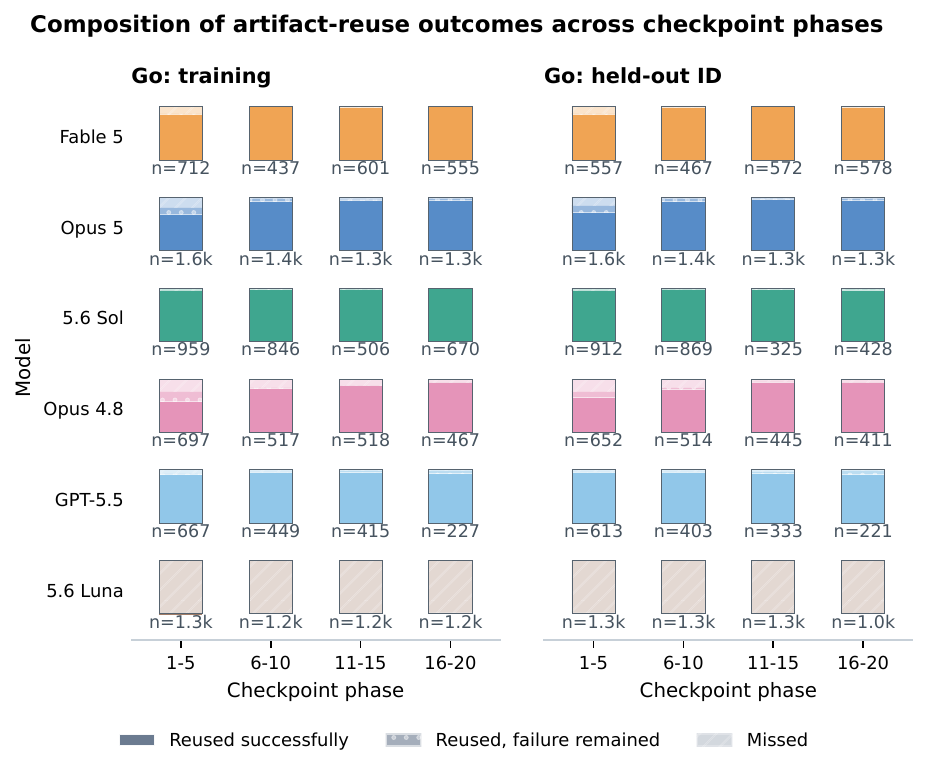}
  \caption{\textbf{Go reuse outcomes over time.} For each model, bars run
  from checkpoints 1--5 to 16--20. Each relevant decision (a move where at least one saved artifact is relevant to choosing or playing it) counts once: reused
  successfully, reused but the failure remained, or missed; $n$ is the number of
  relevant decisions. See Section~\ref{sec:failure-results} and Appendix~\ref{app:gohex-diagnostics}.}
  \label{fig:go-engine-v12-reuse-phases}
\end{figure}

\begin{figure}[H]
  \centering
  \includegraphics[width=0.75\textwidth]{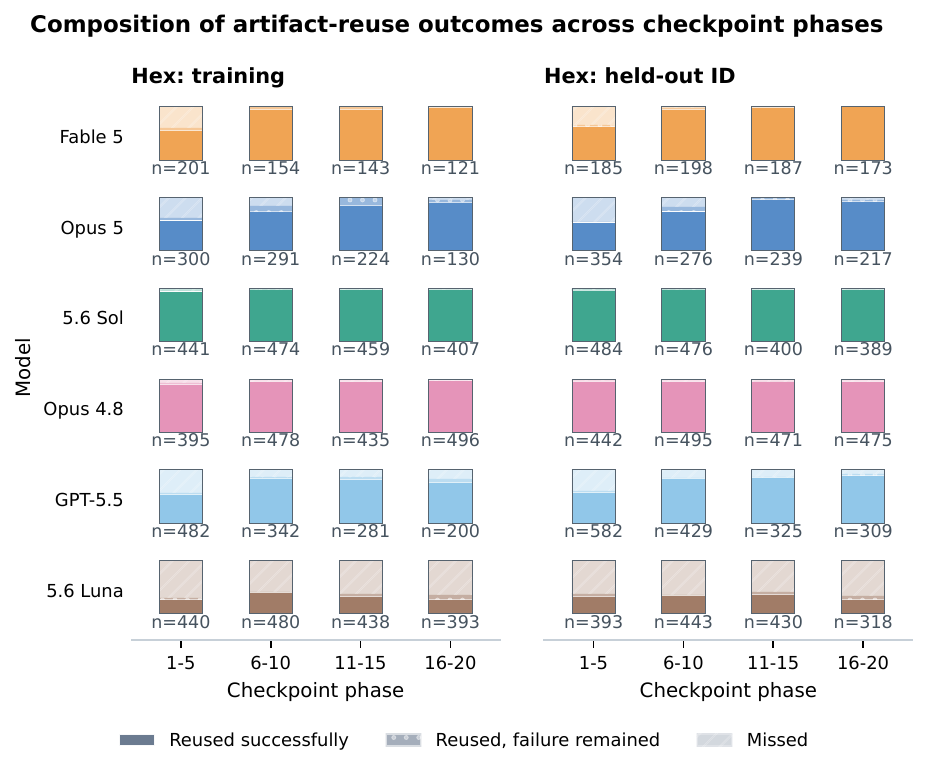}
  \caption{\textbf{Hex reuse outcomes over time.} Same as
  Figure~\ref{fig:go-engine-v12-reuse-phases}. See Section~\ref{sec:failure-results} and Appendix~\ref{app:gohex-diagnostics}.}
  \label{fig:hex-engine-v12-reuse-phases}
\end{figure}

\begin{figure}[H]
  \centering
  \includegraphics[width=0.8\textwidth]{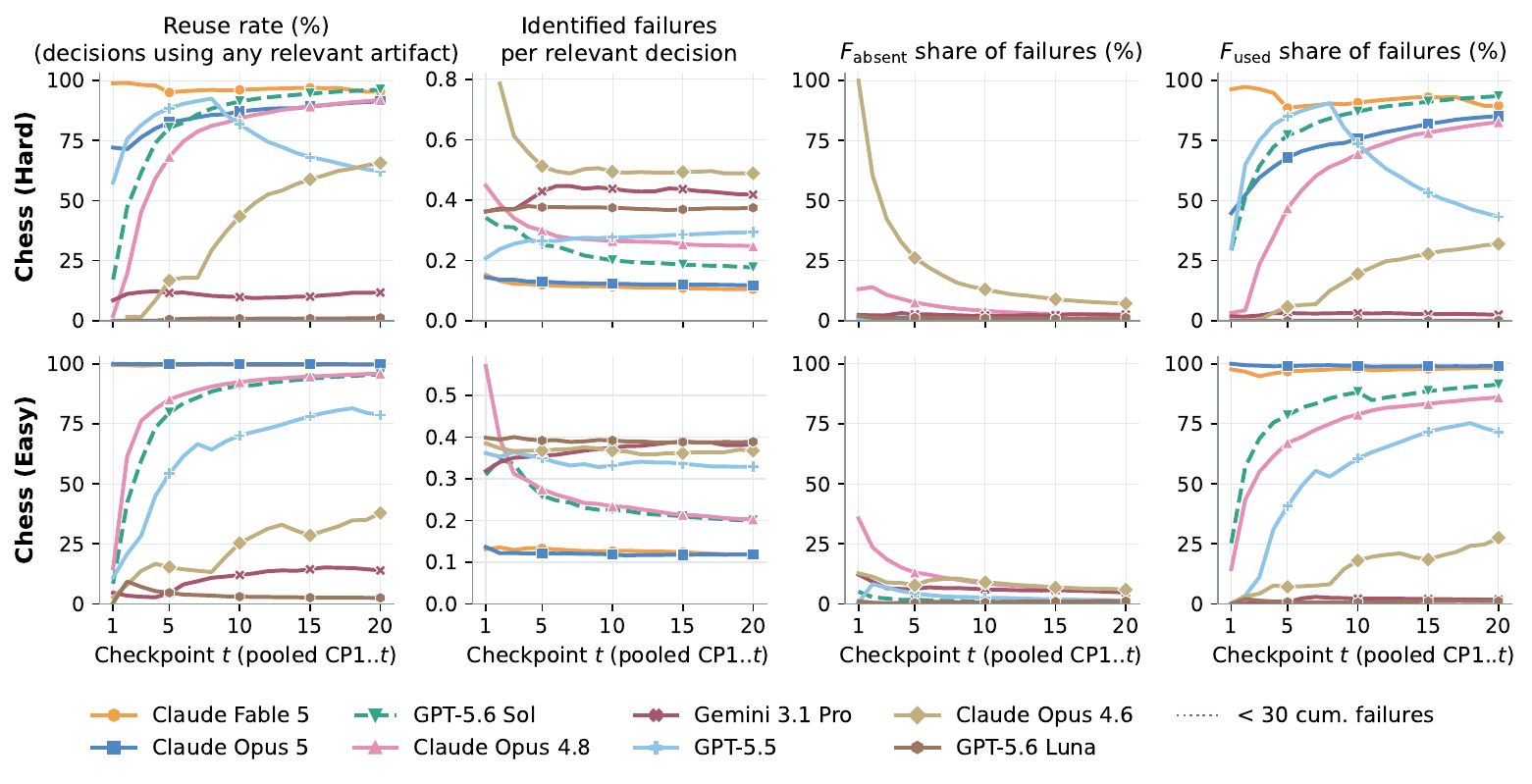}
  \caption{\textbf{Cumulative improvement-loop metrics in chess, all eight models.}
  Same as Figure~\ref{fig:chess-loop-dynamics}, for the Hard (top) and Easy
  (bottom) chess conditions. Each point pools training and held-out ID decisions
  from checkpoint 1 to $t$. Columns: reuse rate; failures per relevant decision (a move where at least one saved artifact is relevant to choosing or playing it);
  $F_{\mathrm{absent}}$; and $F_{\mathrm{used}}$. Dotted: fewer than 30 failures
  so far. See Section~\ref{sec:failure-results} and Appendix~\ref{app:failure-dynamics}.}
  \label{fig:cumulative-loop-metrics-chess}
\end{figure}

\begin{figure}[H]
  \centering
  \includegraphics[width=0.85\textwidth]{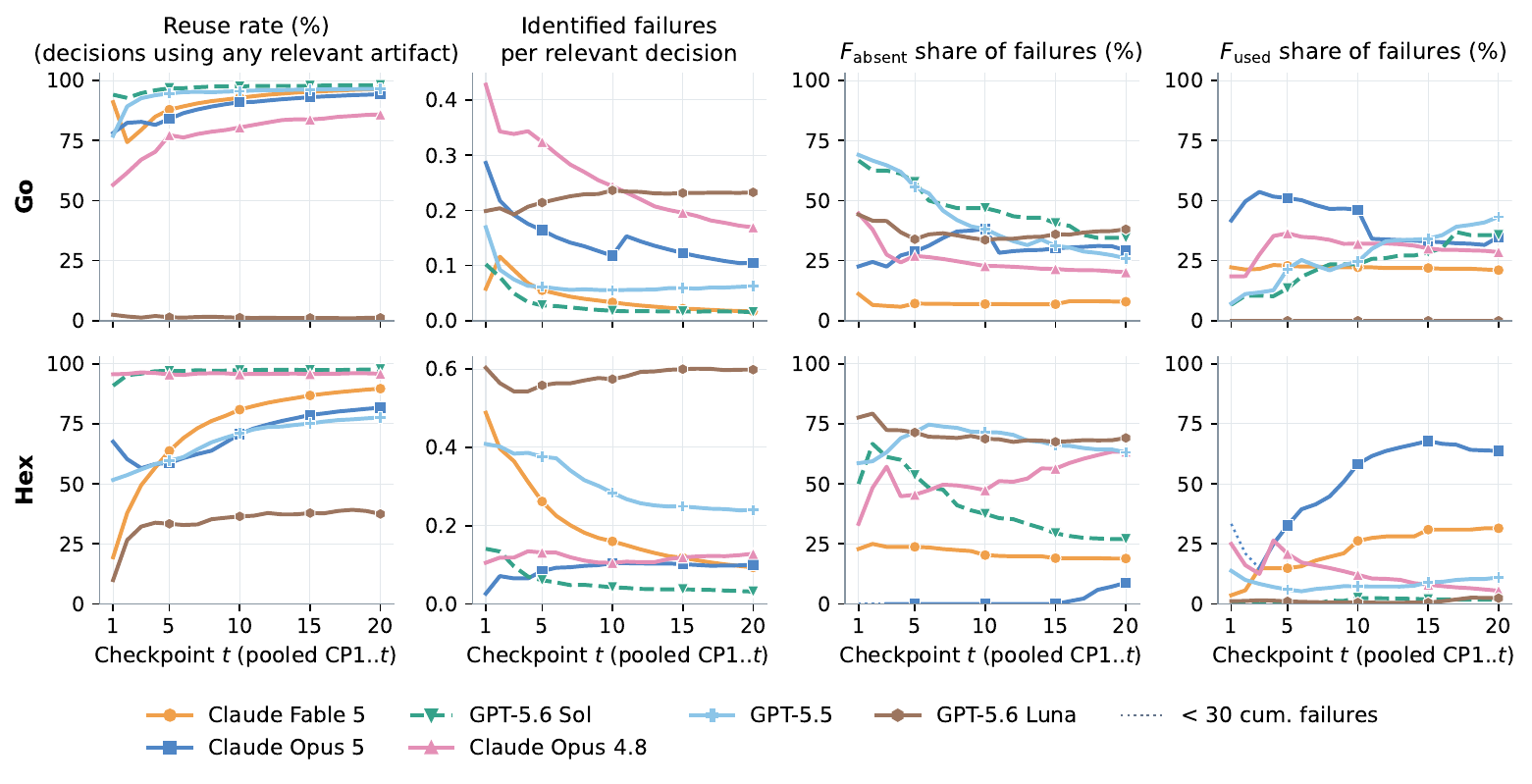}
  \caption{\textbf{Cumulative improvement-loop metrics in Go and Hex.} Same as
  Figure~\ref{fig:chess-loop-dynamics}; each point pools training and held-out ID
  decisions from checkpoint 1 to $t$. Claude Opus~4.6 and Gemini~3.1 Pro are
  analyzed in chess only. Dotted: fewer
  than 30 failures so far. See Section~\ref{sec:failure-results} and Appendix~\ref{app:gohex-diagnostics}.}
  \label{fig:cumulative-loop-metrics-gohex}
\end{figure}

\endgroup

\clearpage
\FloatBarrier
\section{Actor and Reflection Prompts}\label{app:prompts}

Every actor and reflection call receives the system prompt below, then the
game's actor prompt or the reflection prompt, plus a one-line budget message. The
texts are from the harness, with file paths and the validator command
slightly simplified; \texttt{\{...\}} marks a value filled in at
run time and \texttt{[...]} an omission. The prompts are the same for all models
(apart from the tool-call fallback at the end) and for both chess conditions; Go and Hex change only the rules, the game API, the
call budget, and the tool time limit.

\subsection{System prompt (all games, actor and reflection)}
{\footnotesize
\begin{verbatim}
You interact with an environment through a minimal API. Improve your
score over repeated epochs by building reliable infrastructure from your
interactions, feedback, rewards, and failures.

You can create and improve tools such as Python functions, maintain
persistent memory, define strategies for using tools and memory, and
delete inventory that is no longer useful. Decide what to create, use,
revise, retain, or delete.

You have Bash and Python inside a constrained sandbox. Bash can inspect
your persistent inventory and manage files in permitted writable
directories. Python provides the environment API and can execute tools
you create. Python imports and direct filesystem access are unavailable;
use Bash and load_python(path) for permitted files.
\end{verbatim}
}

\subsection{Actor prompts}

The actor has a \texttt{bash} tool (the inventory is read-only and \texttt{/session} is
writable), a persistent \texttt{python} interpreter that exposes the game API and the library's
tools, and a native move-submission tool (\texttt{submit\_move} in chess, \texttt{submit\_action}
in Go and Hex). \texttt{\{color\}} and \texttt{\{actor\_role\}} are the actor's side.

\paragraph{Chess (Hard and Easy).}
{\footnotesize
\begin{verbatim}
A new game has started. You are playing {color}. Persistent
infrastructure is available read-only under /workspace/inventory.
/session is writable and lasts only for this game. Use Bash, Python,
existing infrastructure, newly created temporary tools, or direct
reasoning as you decide. You have at most 64 model calls for this game.

Python exposes board_api as a dictionary with get_fen(),
get_legal_moves(), make_move(move), undo_move(), turn(), and
submit_move(move), for example board_api["get_legal_moves"]().
board_api["submit_move"](move) commits one move in Standard Algebraic
Notation and returns Stockfish's reply and the new position. The native
submit_move tool performs the same action. Submit legal moves to
maximize your score.

Persistent failure-mode and artifact indexes are available under
/workspace/inventory/tracking.
\end{verbatim}
}

\paragraph{Go.}
{\footnotesize
\begin{verbatim}
A new go game has started. You are playing {actor_role}. Persistent
infrastructure is available read-only under /workspace/inventory.
/session is writable and lasts only for this game. Use Bash, Python,
existing infrastructure, newly created temporary tools, or direct
reasoning as you decide. You have at most 32 model calls for this game.

Black uses X and White uses O. Actions are coordinates such as C4 or
pass. Captures, no-suicide, positional superko, two-pass termination,
area scoring, and the displayed komi apply. Final outcomes use Chinese
rules with dead-stone adjudication by the authoritative scorer. The
action cap is 75 total actions. If that cap is reached before two
passes, the scorer adjudicates the current position.

Python exposes game_api as a dictionary with get_state(),
get_legal_actions(), apply_action(action), undo_action(),
current_player(), and submit_action(action).
game_api["submit_action"](action) commits one legal action, lets the
automated opponent respond when the game continues, and returns the new
state. The native submit_action tool performs the same action. Each Bash
or Python call may run for at most 240 seconds. Submit legal actions to
maximize your score.

Persistent failure-mode and artifact indexes are available under
/workspace/inventory/tracking.
\end{verbatim}
}

\paragraph{Hex.} The Hex prompt matches the Go prompt except for the game name and the rules
paragraph.
{\footnotesize
\begin{verbatim}
A new hex game has started. You are playing {actor_role}. [...]

Red uses X and connects north to south. Blue uses O and connects west to
east. Actions are coordinates such as c4. The first completed connection
wins.

[...]
\end{verbatim}
}

\paragraph{Per-call budget message (actor).} This message is appended before every actor model
call (64 calls in chess, 32 in Go and Hex).
{\footnotesize
\begin{verbatim}
Game model-call budget: {used}/{maximum} calls used; {remaining} remain.
\end{verbatim}
}

\paragraph{Tool-timeout recovery (Go and Hex).} If a Bash or Python call exceeds the tool
timeout, the actor gets one final call with only \texttt{get\_state},
\texttt{get\_legal\_actions}, and \texttt{submit\_action} available. This message follows the
budget message.
{\footnotesize
\begin{verbatim}
Your Bash or Python tool timed out. This is your only recovery model
call. You may only inspect the current state, inspect legal actions, and
submit one legal action with the provided native tools. Bash, Python,
and further search are unavailable. Submit a legal action now or the
game is forfeited.
\end{verbatim}
}

\subsection{Reflection prompt}

Reflection has a single \texttt{bash} tool (``Run Bash inside the writable transactional
\texttt{/workspace}.''). The prompt below uses the chess values; Go and Hex use \texttt{300.0}
instead of \texttt{1800.0} seconds. \texttt{/workspace/feedback} contains only the
\emph{training} games of the current checkpoint (moves, engine annotations in chess, traces,
session files), a score summary, and the protocol files (\texttt{SCHEMA.md}, the validator, and
the parent inventory). Held-out games are never copied there.
{\footnotesize
\begin{verbatim}
Game transcripts, annotations, failures, per-game scratch files,
mutation feedback, and your current infrastructure are available under
/workspace. Persistent infrastructure is under /workspace/inventory. On
a retry, prior reflection logs are available under
/workspace/feedback/reflection_attempts. Improve the infrastructure for
future games.

You can create or modify Python tools, create or modify persistent
memory, create or modify strategies for using tools and memory, and
delete inventory that is no longer useful. Use the Bash tool to inspect
and edit /workspace/inventory. Invoke Bash directly rather than
describing or printing a tool call. Each Bash call may run for at most
1800.0 seconds. You have at most 50 reflection model calls and will be
told how many remain. Finish early when the work is complete. The
inventory is validated after the final call.

Maintain failure_modes.json and artifacts.json under
/workspace/inventory/tracking while inspecting the canonical
training trajectories under /workspace/feedback. Reuse a failure-mode ID
when the causal problem is the same; create a new ID only for a
materially different problem. Map failures to exact training game IDs. A
trajectory need not be mapped when you find no failure. Preserve
mitigated and merged modes rather than deleting them.

Maintain stable artifact IDs for persistent tools, memory, strategies,
instructions, and other infrastructure. Keep an ID when revising the
same capability, but create a new ID for a materially different
capability rather than repurposing an old ID. Record paths, executable
entrypoints, addressed failure modes, and active or retired status.
Every functional inventory file you create, modify, or retire must
belong to an artifact addressing at least one failure mode. Retired
artifacts must not leave implementation files in the active inventory.

Use these exact JSON shapes. `failure_modes.json` is `{"schema_version":
"failure_modes.v1","failure_modes":[{"id":"FM-0001","description":"...",
"status":"active","trajectory_ids":["checkpoint_000_train_s1_white"]}]}`
. Status is `active`, `mitigated`, or `merged`; only a merged entry adds
`"merged_into":"FM-..."`. `artifacts.json` is `{"schema_version":"artifa
cts.v1","artifacts":[{"id":"AR-0001","description":"...","kind":"tool","
paths":["tools/example.py"],"entrypoints":["tools/example.py:choose_move
"],"failure_mode_ids":["FM-0001"],"status":"active"}]}`. Kind is `tool`,
`memory`, `strategy`, `instruction`, or `other`; status is `active` or
`retired`; non-executable artifacts use an empty `entrypoints` list. The
same schemas and transition rules are in
/workspace/feedback/protocol/SCHEMA.md.

Before finishing, run the protocol validator
/workspace/feedback/protocol/validate_tracking.py and correct
structural errors before the final response.
\end{verbatim}
}

\paragraph{Per-call budget message (reflection).}
{\footnotesize
\begin{verbatim}
Reflection budget: {used}/{maximum} calls used; {remaining} remain.
\end{verbatim}
}

\paragraph{Final reflection call.} This replaces the budget message on the last permitted call.
{\footnotesize
\begin{verbatim}
Reflection budget: {used}/{maximum} calls used; {remaining} remains.
Final reflection call {turn}/{maximum}. Given the evidence so far, make
the best persistent infrastructure edits now. Invoke the Bash tool
directly for edits under /workspace/inventory rather than describing or
printing a tool call. The inventory is validated after this call.

Synchronize failure_modes.json and artifacts.json with the final
inventory, then run the protocol validator before finishing.
\end{verbatim}
}

\paragraph{Retry addendum.} If a reflection attempt is rejected (for example, the new library
fails to load), the harness starts a new attempt with this text appended (up to 6 attempts in
chess, 3 in Go and Hex). \texttt{\{latest\_error\}} holds the rejection message; a rendered
example from a Chess~(Hard) run follows the template.
{\footnotesize
\begin{verbatim}
This is a retry after a rejected reflection attempt. Before editing,
read /workspace/feedback/reflection_retry_history.json and the prior
logs under /workspace/feedback/reflection_attempts. Resolve the recorded
validation error instead of repeating the rejected change.

Latest rejection: {latest_error}
\end{verbatim}
}
\noindent Rendered example of the last line:
{\footnotesize
\begin{verbatim}
Latest rejection: ValueError: candidate infrastructure failed smoke
validation: persistent tool load failed: NameError: name 'object' is not
defined
\end{verbatim}
}

\paragraph{Tool-call fallback.} Most Go and Hex runs of models other than Claude Fable~5 go
through an inference gateway, which appends this sentence to the system prompt.
{\footnotesize
\begin{verbatim}
Use structured tool calling whenever you call a tool. If structured tool
calling is unavailable, end your response with FINAL_TOOL_CALL_JSON: on
its own line, followed by exactly one JSON object in the form
{"functionCall":{"name":"tool_name","args":{}}}. Anything after that
marker is treated as a real tool call, so never use the marker in
examples.
\end{verbatim}
}

\end{document}